\documentclass[journal]{IEEEtran}
\usepackage{amsmath,amsopn,amssymb}
\usepackage{graphicx,xspace,color,soul}
\usepackage{epsfig,subfigure}
\usepackage{longtable,multirow}
\usepackage{array,float}
\usepackage{cite}
\usepackage{algorithmic}
\usepackage[linesnumbered, ruled]{algorithm2e}
\usepackage{bm,epstopdf}
\usepackage{tabularx}

\renewcommand{\vec}[1]{\mathbf{#1}}

\DeclareMathOperator*{\argmax}{arg\,max}
\DeclareMathOperator*{\argmin}{arg\,min}

\newcolumntype{Y}{>{\centering\arraybackslash}X}

\def\f{{\mathbf f}}
\def\v{{\mathbf v}}
\def\M{{\mathbf M}}
\def\Q{{\mathbf Q}}

\def\bLambda{{\boldsymbol \Lambda}}

\def\ie{{\textit{i.e.}}}

\begin{document}


\title{Feature Graph Learning for 3D Point Cloud Denoising}

\author{
    Wei~Hu,~\IEEEmembership{Member,~IEEE,}
	Xiang~Gao,~\IEEEmembership{Student Member,~IEEE,}
	Gene Cheung,~\IEEEmembership{Senior Member,~IEEE,}
	and~Zongming~Guo,~\IEEEmembership{Member,~IEEE}
\thanks{W. Hu (e-mail: forhuwei@pku.edu.cn), X. Gao (e-mail: gyshgx868@pku.edu.cn), and Z. Guo (e-mail: guozongming@pku.edu.cn) are with Wangxuan Institute of Computer Technology, Peking University, No. 128 Zhongguancun North Street, Beijing, China. G. Cheung (e-mail: genec@yorku.ca) is with York University, 4700 Keele Street
Toronto, Ontario, Canada.}
}


\maketitle

\begin{abstract}
Identifying an appropriate underlying graph kernel that reflects pairwise similarities is critical in many recent graph spectral signal restoration schemes, including image denoising, dequantization, and contrast enhancement. 
Existing graph learning algorithms compute the most likely entries of a properly defined graph Laplacian matrix $\mathbf{L}$, 
but require a large number of signal observations $\mathbf{z}$'s for a stable estimate.
In this work, we assume instead the availability of a relevant feature vector $\mathbf{f}_i$ per node $i$, from which we compute an optimal feature graph via optimization of a feature metric.
Specifically, we alternately optimize the diagonal and off-diagonal entries of a Mahalanobis distance matrix $\mathbf{M}$ by minimizing the graph Laplacian regularizer (GLR) $\mathbf{z}^{\top} \mathbf{L} \mathbf{z}$, where edge weight is $w_{i,j} = \exp \{ - (\mathbf{f}_i - \mathbf{f}_j)^{\top} \mathbf{M} (\mathbf{f}_i - \mathbf{f}_j) \}$, given a single observation $\mathbf{z}$. 
We optimize diagonal entries via proximal gradient (PG), where we constrain $\mathbf{M}$ to be positive definite (PD) via linear inequalities derived from the Gershgorin circle theorem. 
To optimize off-diagonal entries, we design a block descent algorithm that iteratively optimizes one row and column of $\mathbf{M}$.  
To keep $\mathbf{M}$ PD, we constrain the Schur complement of sub-matrix $\mathbf{M}_{2,2}$ of $\mathbf{M}$ to be PD when optimizing via PG. 
Our algorithm mitigates full eigen-decomposition of $\mathbf{M}$, thus ensuring fast computation speed even when feature vector $\mathbf{f}_i$ has high dimension. 
To validate its usefulness, we apply our feature graph learning algorithm to the problem of 3D point cloud denoising, resulting in state-of-the-art performance compared to competing schemes in extensive experiments.


\end{abstract}


\section{Introduction}
\label{sec:intro}

Graphs are flexible mathematical structures modeling pairwise relations between data entities, such as brain networks, social networks, computer networks and transportation networks. 
Nodes in a graph represent data collecting entities (e.g., users in a social group or sensors in a wireless network) while edges connecting nodes describe pairwise affinities.
A scalar (weight) is often assigned to each edge, which reflects the degree of pairwise similarity between two nodes. 
In settings where the graph is not readily available, it is critical to first identify an appropriate underlying graph kernel---a process commonly called \textit{graph learning} \cite{Giannakis18graph,dong19spm,Mateos19spm}---before it is used for many recent graph spectral signal restoration schemes, including image denoising, dequantization, deblurring, and contrast enhancement \cite{pang2017graph,liu2016random,bai18,liu19contrast}.

Existing graph learning methods\footnote{We focus on learning of \textit{undirected} graphs when one or fewer signal observation is available in this paper, while learning of \textit{directed} graphs \cite{Mateos19spm} is left for future work.} can be roughly classified into two main categories: statistical methods and graph spectral methods. 
The common assumption behind statistical methods is that \textit{multiple} data observations generated from the same probability model are available to determine the model parameters, which describe an underlying graph. 
Graph learning is then essentially the problem of estimating the inverse covariance or precision matrix given sufficient empirical data, with the addition of some prior topological information (e.g., sparsity) \cite{meinshausen2006high, banerjee2008model,friedman2008sparse,
dempster1972covariance,ravikumar2011high,cai2011constrained,mazumder2012graphical,mazumder2012exact,hsieh2013big,grechkin2015pathway}. 
However, the requirement of multiple signal observations is not applicable to scenarios where only a \textit{single} observed signal is available.

On the other hand, graph spectral learning methods offer an alternative (or additional) \textit{signal representation} perspective, where observed signals are assumed to lie in a low-dimensional subspace spanned by the low frequency components of the underlying graph topologies \cite{lake2010discovering, kalofolias2016learn,sardellitti2016graph, dong2016learning, segarra2017network,pasdeloup2017characterization,egilmez2017graph,jiang19cvpr}. 
Specifically, the frequency components are eigenvectors of a chosen variational operator on graphs like the adjacency matrix or the graph Laplacian matrix \cite{chung97}. 
This ``low-pass" spectral assumption translates to additional constraints during graph learning, potentially leading to more accurate graph estimates when there are few signal observations. 


Extending on these previous works \cite{dong2016learning,egilmez2017graph,yang2018apsipa}, in this paper we study spectral graph learning when the number of signal observations is extremely small---just one observation or even fewer (i.e., partial observation of one signal). 
This is typically the case for image restoration applications with non-stationary statistics \cite{pang2017graph,liu2016random,bai18,liu19contrast}, where the underlying graph for a target image patch needs to be estimated for graph spectral processing given just one noisy and/or partial patch observation. 
To ease the ill-posedness of the problem, we assume the availability of a relevant feature vector $\mathbf{f}_i$ per node $i$, $\mathbf{f}_i \in \mathbb{R}^K$ (e.g., the color pixel intensities), and that an edge weight is an inverse function of the feature distance (i.e., larger the inter-node distance, smaller the edge weight).
Many previous graph constructions including bilateral filter \cite{tomasi1998bilateral,shuman13spm,hu14tip,pang2017graph,yang2018apsipa} implicitly assume some notion of feature distance when assigning edge weights; our work is a more formal study of feature metric learning in a rigorous mathematical setting.

Specifically, we assume an edge weight $w_{i,j} = \exp \{ - (\mathbf{f}_i - \mathbf{f}_j)^{\top} \mathbf{M} (\mathbf{f}_i - \mathbf{f}_j) \}$, where $\mathbf{M} \in \mathbb{R}^{K \times K}$ is the  \textit{Mahalanobis distance} metric matrix \cite{mahalanobis1936} given feature difference $\mathbf{f}_i - \mathbf{f}_j$ for the two connected nodes $i$ and $j$.
Given a single observation $\mathbf{z} \in \mathbb{R}^N$, we seek to minimize the \textit{Graph Laplacian Regularizer} (GLR) \cite{pang2017graph} $\mathbf{z}^{\top} \mathbf{L}(\mathbf{M}) \mathbf{z}$ using $\mathbf{M}$, which measures the smoothness of the signal $\mathbf{z}$ with respect to the graph Laplacian matrix $\mathbf{L}$.
Note that because the feature dimension $K$ is often much smaller than signal dimension $N$, variable $\mathbf{M}$ has $K^2$ entries only, while in general Laplacian matrix $\mathbf{L}$ has $N^2$ entries, resulting in more stable parameter estimation given the same observed data.

To find the optimal $\mathbf{M}$, we alternately optimize the diagonal and off-diagonal entries. 
When optimizing diagonal entries, we derive linear inequalities based on the \textit{Gershgorin circle theorem} (GCT) \cite{gervsgorin1931uber} to keep $\mathbf{M}$ positive definite (PD) and then employ proximal gradient (PG) descent \cite{Parikh31} to acquire a solution.   
When optimizing off-diagonal entries, we design a block descent algorithm that iteratively optimizes one row and column of $\mathbf{M}$ until convergence.
To keep $\mathbf{M}$ PD, we constrain the Schur complement of $\mathbf{M}$ to be PD according to the \textit{Haynsworth inertia additivity} \cite{haynsworth1968determination}, which is enforced using PG descent. 
Further, we relax the PD constraint via vector-norm bounding to avoid matrix inverse computation during the optimization. 
Our algorithm also mitigates full eigen-decomposition of $\mathbf{M}$, thus ensuring fast computation speed even when the feature dimension $K$ is large. 

To validate the usefulness of our proposed feature graph learning algorithm, we apply it to the problem of 3D point cloud denoising. 
Point clouds provide efficient representation for arbitrarily-shaped objects, which consist of a set of irregularly-spaced points. 
The maturity of depth sensing and 3D laser scanning techniques\footnote{Commercial products include Microsoft Kinect, LiDAR, Intel RealSense, etc. } enables convenient acquisition of 3D point clouds, which have a variety of applications such as 3D immersive tele-presence, navigation for autonomous vehicles, free-viewpoint rendering, and heritage reconstruction \cite{tulvan16}. 
However, point clouds are often perturbed by noise, which comes from hardware, software or other causes.  

To denoise point clouds, we first assume that local patches are \textit{self-similar} \cite{buades2005non, rosma13} and connect corresponding points into a graph.
Assuming a \textit{first-order intrinsic Gaussian Markov random field} (IGMRF) model \cite{rue2005gaussian}, we pose a \textit{Maximum a Posteriori} (MAP) estimation problem with GLR as signal prior. 
Interpreting the precision matrix in the IGMRF model as a graph Laplacian, we employ the feature graph learning to optimize edge weights, where for each point we employ 3D coordinates and surface normals as relevant features.
Finally, we optimize the point cloud and the underlying graph alternately until convergence. 
Extensive experiments show that we achieve state-of-the-art performance compared to competing methods \cite{mattei,sarkar2018structured,zeng18arxiv}.

To summarize, the main contributions of our works are:
\begin{enumerate}
    \item To identify an appropriate underlying graph given a single signal observation $\mathbf{z}$, we formulate a feature graph learning problem by minimizing the GLR $\mathbf{z}^{\top} \mathbf{L}(\mathbf{M}) \mathbf{z}$ using the Mahalanobis distance metric matrix $\mathbf{M}$ as variable, assuming feature vector per node is available. 
    \item We develop a fast block descent algorithm to optimize the feature metric matrix $\mathbf{M}$, while keeping $\mathbf{M}$ positive definite and mitigating full matrix eigen-decomposition and large matrix inverse; 
    \item We employ feature graph learning to 3D point cloud denoising, where the graph for each set of self-similar patches is computed from 3D coordinates and surface normals as features, resulting in superior denoising performance. 
\end{enumerate}

The paper is organized as follows. 
We first review previous works on graph learning and point cloud denoising in Section~\ref{sec:related}. 
Then we introduce basic concepts in graph spectral processing in Section~\ref{sec:graph}. 
In Section~\ref{sec:learning}, we describe the proposed problem formulation and algorithm development for feature graph learning. 
We then apply to the problem of point cloud denoising in Section~\ref{sec:formulation}. 
Finally, experimental results and conclusions are presented in Section~\ref{sec:results} and \ref{sec:conclude}, respectively.

\section{Related Work}
\label{sec:related}
We overview previous works on graph learning and point cloud denoising in order. 

\subsection{Graph Learning}

Previous graph learning methods can be divided into two main categories: statistical methods and graph spectral methods.

\textbf{Statistical methods:} 
In graphical models including Markov random fields \cite{rue2005gaussian} and Bayesian networks \cite{jensen1996introduction}, edges in the graph encode conditional dependencies among random variables represented as nodes. 
Learning the graph structure amounts to learning the inverse covariance or precision matrix for such models. 
Dempster \cite{dempster1972covariance} proposed to introduce zero entries in inverse covariance matrices for simplified covariance estimation. 
The estimation of a sparse inverse covariance matrix was then studied in several works \cite{ravikumar2011high,cai2011constrained,meinshausen2006high}.
Friedman \textit{et al.} formulated sparse inverse covariance estimation with a regularization framework and developed the Graphical Lasso algorithm to address the regularized optimization problem \cite{friedman2008sparse}.
Some algorithmic extensions of the Graphical Lasso are presented in \cite{banerjee2008model,mazumder2012graphical}, and a few computationally efficient variations are discussed in \cite{mazumder2012exact,hsieh2013big,grechkin2015pathway}. 
However, inverse covariance estimation methods assume many observations of a graphical model, which is not applicable for many imaging applications.

\textbf{Graph spectral methods:} 
The key idea is to enforce low frequency representation of observed signals as well as constraints for a valid graph Laplacian matrix. 
Tenenbaum \textit{et al.} \cite{lake2010discovering} proposed to learn combinatorial graph Laplacians using a proposed sparse model. 
A regression framework was presented in \cite{hu2013graph} to learn a graph Laplacian matrix based on a fitness metric between the signals and the graph, which essentially evaluates the smoothness of the signals on the graph. 
Dong \textit{et al.} \cite{dong2016learning} and Kalofolias \cite{kalofolias2016learn} proposed to learn Laplacian matrices from the smoothness prior of the graph signal. 
Egilmez \textit{et al.} \cite{egilmez2016arxiv} proposed graph learning under pre-defined graph structural and graph Laplacian constraints.
Yang \textit{et al.} \cite{yang2018apsipa} computed optimal feature weights in a similarity graph given a restored binary classifier signal. 
This is an earlier version of our feature metric learning, but restricts the search space only to diagonal matrices, which limits its effectiveness.

Orthogonally, some studies focus on inferring graph topologies from signals that are diffused on a graph over time. 
A fitness metric similar to the regression framework was employed in \cite{daitch2009fitting} to learn a valid graph topology. 
In particular, Segarra \textit{et al.} \cite{segarra2017network} and Pasdeloup \textit{et al.} \cite{pasdeloup2017characterization} focused on learning graph shift/diffusion operators (such as adjacency matrices) from a set of diffused graph signals.
Sardellitti \textit{et al.} \cite{sardellitti2016graph} proposed to learn the graph topology from data under the assumption of band-limited signals, which corresponded to signals with clustering properties.
Nonetheless, this class of methods also assume a large number of signal observations for a stable estimate.


\subsection{Point Cloud Denoising}

Point cloud denoising methods mainly include Moving Least Squares (MLS) based methods, Locally Optimal Projection (LOP) based methods, sparsity based methods, non-local based methods and graph-based methods. 

\textbf{MLS-based methods:}~~~MLS-based methods approximate the point cloud with a smooth surface and then project the points of the point cloud onto the fitted surface. 
\cite{alexa01} used the MLS projection operator to calculate the optimal MLS surface of the point cloud, and moved the points around the surface to the MLS surface. 
\cite{guennebaud07} proposed a MLS-based spherical fitting denoising method (APSS). 
Compared with the aforementioned MLS projection-based algorithm, this method improved the stability at low sampling rate and high curvature. 
\cite{ouml10} proposed an algorithm based on improved MLS and local kernel regression to smooth the point cloud surface (RIMLS). 
However, these MLS-based methods are often sensitive to outliers.

\textbf{LOP-based methods:}~~~The widely known LOP \cite{lipman07} aimed to produce a set of points to represent the underlying surface while enforcing a uniform distribution over the point cloud. 
Weighted LOP (WLOP) \cite{wlop09} provided a more uniformly distributed output than LOP by adapting a repulse term to the local density. 
Further, anisotropic WLOP (AWLOP) \cite{awlop13} modified WLOP with an anisotropic weighting function in order to preserve sharp features better. 
Nevertheless, LOP-based methods often suffer from over-smoothing.

\textbf{Sparsity based methods:}~~~These methods are based on sparse representation theory \cite{xu2015survey}, and generally involve two phases. 
In the first phase, the sparse reconstruction of the cloud normals is obtained by solving the global minimization problem of sparse regularization. 
In \cite{avron10} $l_1$ regularization was adopted, while \cite{sun15} used $l_0$ regularization to seek more characteristic sparsity.
In the second phase, each point position is updated by solving global $l_1$ (or $l_0$) minimization problem based on the reconstructed normals and local planarity hypothesis. 
The recently proposed method called \textit{Moving Robust Principal Components Analysis} (MRPCA) \cite{mattei} used weighted $l_1$ minimization of the point deviations from the local reference plane to preserve sharp features.
However, when the noise level is high, over-smoothing or over-sharpening tends to occur \cite{sun15}.

\textbf{Non-local based methods:}~~~These approaches exploit the non-local similarities among patches in a point cloud. 
In \cite{deschaud10}, an extended \textit{non-local denoising} (NLD) algorithm was introduced to process point clouds, where the neighborhood of each point was described by the polynomial coefficients of the local MLS surface to compute point similarity.
\cite{digne12} and \cite{guillemot12} applied a scale space scheme and non-local means denoising algorithm. 
\cite{rosma13} extended the BM3D \cite{dabov2007image} algorithm to point cloud denoising, searched similar patches globally via Iterative Closest Point (ICP) \cite{icp} algorithm, and then combined them into a collaborative group for denoising. 
\cite{sarkar2018structured} utilized patch self-similarity and optimized for a low rank (LR) dictionary representation of the extracted patches to smooth 3D patches.
However, the computational complexity of such methods is often high due to the global search.

\textbf{Graph-based methods:}~~~This class of methods interpret a point cloud as a signal on a graph, and perform denoising via chosen graph filters. 
In \cite{Schoenenberger2015Graph}, the input point cloud was represented as a signal on a $k$-nearest-neighbor graph and then denoised via a convex optimization problem regularized by the gradient of the point cloud on the graph. 
In \cite{dinesh18arxiv}, a reweighted graph Laplacian regularizer for surface normals was designed, with a general $l_p$-norm fidelity term that modeled two types of additive noise. 
Moreover, they established a linear relationship between normals and 3D point coordinates via bipartite graph approximation for ease of optimization. 
\cite{zeng18arxiv} proposed graph Laplacian regularization (GLR) of a low dimensional manifold model (LDMM), and sought self-similar patches to denoise them simultaneously. 
Instead of directly smoothing the coordinates or normals of 3D points, \cite{duan2018weighted} estimated a local tangent plane at each
3D point based on a graph and then reconstructed each 3D point by weighted averaging of its projections on multiple tangent planes. 

Our proposed approach belongs to the family of graph-based methods. 
The key difference is that edge weights in our graph are not pre-defined with hand-crafted parameters, but optimized rigorously via feature metric learning given available signal(s) assumed to be smooth with respect to the graph. 


\section{Background on Spectral Graph Theory}
\label{sec:graph}
\begin{table}[t]
\centering
\caption{List of abbreviations and their descriptions.}
\begin{tabular}{|c|c|}
    \hline
    \textbf{Abbreviation} & \textbf{Description} \\ \hline
    GLR  & Graph Laplacian Regularizer  \\
    PD   & Positive Definite \\
    PSD   & Positive Semi-Definite \\
    PG   & Proximal Gradient \\
    IGMRF & Intrinsic Gaussian Markov Random Fields \\
    MAP  & Maximum a Posteriori \\
    MLS  & Moving Least Squares \\
    RIMLS & Robust Implicit MLS \\
    APSS & Algebraic Point Set Surfaces \\
    LOP & Locally Optimal Projection \\
    WLOP & Weighted LOP \\
    AWLOP & Anisotropic WLOP \\
    NLD & Non-Local Denoising \\
    LR & Low Rank \\ \hline
    \end{tabular}
\label{table:abbr}
\end{table}

We first review basic concepts in spectral graph theory \cite{chung97} that are essential in our feature graph learning and point cloud denoising algorithms. 

\subsection{Graph and Graph Laplacian}
\label{subsec:laplacian}

We consider an undirected graph $ \mathcal{G}=\{\mathcal{V},\mathcal{E}, \mathbf{W}\} $ composed of a node set $ \mathcal{V} $ of cardinality $|\mathcal{V}|=N$, an edge set $ \mathcal{E} $ connecting nodes, and a weighted adjacency matrix $\mathbf{W}$. Each edge $(i,j) \in \mathcal{E}$ is associated with a non-negative weight $w_{i,j}$ which reflects the degree of similarity between nodes $i$ and $j$.  

Among different variants of Laplacian matrices, in this paper we employ the \textit{combinatorial graph Laplacian} \cite{shen10pcs,hu12icip,hu14tip} defined as $ \mathbf{L}:=\mathbf{D}-\mathbf{W} $, where $ \mathbf{D} $ is the \textit{degree matrix}---a diagonal matrix where $ d_{i,i} = \sum_{j=1}^N w_{i,j} $. 






\subsection{Graph Laplacian Regularizer} 
\label{subsec:gssp}

Graph signal refers to data that resides on the nodes of a graph, such as functionality of regions on a neural network and temperatures on a sensor network. 


A graph signal $ \vec{z} \in \mathbb{R}^N $ defined on a graph $ \mathcal{G} $ is \textit{smooth} with respect to $ \mathcal{G} $ \cite{spielman2007spectral} if  
\begin{equation}
	\vec{z}^{\top} \mathbf{L} \vec{z} =\sum_{i=1}^{N} \sum_{j=1}^{N} w_{i,j}(z_i - z_j)^2 < \epsilon,
	\label{eq:prior}
\end{equation}
where $ \epsilon $ is a small positive scalar. 
To satisfy (\ref{eq:prior}), connected node pair $ z_i $ and $ z_j $ must be similar for a large edge weight $ w_{i,j} $; for a small $ w_{i,j} $, $z_i$ and $z_j$ can differ significantly. 
Hence, (\ref{eq:prior}) forces $ \vec{z} $ to adapt to the topology of $ \mathcal{G} $, and is commonly called the \textit{graph Laplacian Regularizer} (GLR) \cite{shuman13spm,pang2017graph}.
This prior also has a frequency interpretation:
\begin{align}
\vec{z}^{\top} \mathbf{L} \vec{z} =
\sum_k \lambda_k \alpha_k^2
\end{align}
where $\lambda_k$ is the $k$-th eigenvalue of $\mathbf{L}$ and is commonly interpreted as the $k$-th graph frequency, and $\alpha_k = \mathbf{v}_k^{\top} \mathbf{z}$ is the inner-product between the corresponding $k$-th eigenvector and signal $\mathbf{z}$. 
In other words, $\alpha_k^2$ is the energy in the $k$-th graph frequency for signal $\mathbf{z}$. 
Thus, a small $\vec{z}^{\top} \mathbf{L} \vec{z}$ means that most signal energies are in the low graph frequencies, or $\mathbf{z}$ is roughly low-pass.


\subsection{Signal-Dependent Graph Laplacian Regularizer}
\label{subsec:reweighted}

In the aforementioned GLR, the graph Laplacian $\mathbf{L}$ is fixed, which does not promote reconstruction of the target signal with discontinuities if the corresponding edge weights are not very small. 
It is thus extended to \textit{signal-dependent} GLR in \cite{liu2016random,pang2017graph,bai18} by considering $\mathbf{L}(\vec{z})$ as a function of the graph signal $\vec{z}$. 
Specifically, an edge weight $w_{i,j}(z_i,z_j)$ is an inverse function of the inter-node pixel intensity difference, e.g., $w_{i,j}(z_i, z_j) = \exp \{ - (z_i - z_j)^2/\sigma^2 \}$. 
The reweighted prior is defined as    
\begin{equation}
    \vec{z}^{\top} \mathbf{L}(\vec{z}) \vec{z} = \sum\limits_{i \sim j} w_{i,j}(z_i, z_j) \cdot (z_j - z_i)^2,
    \label{eq:rgssp}
\end{equation}
where $w_{i,j}(z_i, z_j)$ is the $(i,j)$-th element of the corresponding adjacency matrix $\mathbf{W}$.

It has been shown in \cite{bai18,pang2017graph,liu2016random} that minimizing the signal-dependent GLR iteratively can promote piecewise smoothness (PWS) in the reconstructed graph signal $\vec{z}$, assuming that edge weights are appropriately initialized. 
In our more general setting, given a feature vector per node---which may include the signal intensity as one feature---our work can be considered a general case that includes signal-dependent GLR as a special case, where we compute the best feature graph via an optimization of the Mahalanobis distance metric.     

\section{Feature Metric Learning}
\label{sec:learning}
\subsection{Problem Formulation}
\label{subsec:formulation}
Conceptually, an edge weight $w_{i,j}$ reflects the similarity between samples at nodes $i$ and $j$; specifically, using the commonly used Gaussian kernel \cite{shuman13spm}, edge weight $w_{i,j} = \exp\left\{ - \delta_{i,j} \right\}$, where $\delta_{i,j}$ denotes the estimated feature distance between samples $i$ and $j$.
One advantage of the Gaussian kernel is that edge weight $w_{i,j}$ is in range $[0,1]$, ensuring the resulting combinatorial graph Laplacian matrix $\mathbf{L}$ to be \textit{positive semi-definite} (PSD) \cite{cheung2018graph}.

The feature distance between two samples essentially measures the inter-sample similarity.    
As one well-known example of feature distance, consider the \textit{bilateral filter} in image denoising \cite{tomasi1998bilateral} 
that employs pixel intensities $x_i$ and pixel locations $\mathbf{l}_i$ as relevant features to compute $\delta_{i,j}$, namely,
\begin{align}
\delta_{i,j} = \frac{(x_i - x_j)^2}{\sigma_x^2} + \frac{\|\mathbf{l}_i - \mathbf{l}_j\|_2^2}{\sigma_l^2},
\label{eq:bf}
\end{align}
where $\sigma_x$ and $\sigma_l$ are parameters. 
Defining $\mathbf{f}_i = [x_i ~~ \mathbf{l}_i]^{\top}$, we can rewrite \eqref{eq:bf} in matrix form as:
\begin{align}
\delta_{i,j} = \left( \mathbf{f}_i - \mathbf{f}_j \right)^{\top} 
\left[ \begin{array}{cc}
1/\sigma_x^2 & 0 \\
0 & 1/\sigma_l^2
\end{array}
\right]
\left( \mathbf{f}_i - \mathbf{f}_j \right).
\end{align}
\noindent
\cite{tomasi1998bilateral} shows that with appropriate parameters $\sigma_x$ and $\sigma_l$, the bilateral filter can achieve very good edge-preserving image denoising performance. 
How to best determine $\sigma_x$ and $\sigma_l$, however, was left unanswered.

More generally, associated with each sample $i$ is a length-$K$ vector of relevant features, and our goal is to compute an optimal \textit{Mahalanobis distance} for the given features: 
\begin{equation}
    \delta_{i,j} = (\mathbf{f}_i-\mathbf{f}_j)^{\top} \mathbf{M} (\mathbf{f}_i-\mathbf{f}_j),
    \label{eq:graph_weight}
\end{equation}
where $\mathbf{M} \in \mathbb{R}^{K \times K}$ is a \textit{positive definite} (PD) matrix\footnote{PD is desirable, so that if $\mathbf{f}_i - \mathbf{f}_j \neq \mathbf{0}$, then $(\mathbf{f}_i-\mathbf{f}_j)^{\top} \mathbf{M} (\mathbf{f}_i-\mathbf{f}_j) > 0$.}. 
%
%
As a special case, when $\mathbf{M}$ is a diagonal matrix with strictly positive diagonal entries, the definition in (\ref{eq:graph_weight}) defaults to that in \cite{yang2018apsipa}. 
Diagonal $\mathbf{M}$ can capture the relative importance of individual features when computing $\delta_{i,j}$, but fails to capture possible cross-correlation among features, and thus is sub-optimal in the general case.    

\subsubsection{Importance of Off-diagonal Terms in $\mathbf{M}$}
For completeness, we illustrate the importance of off-diagonal terms via the following analysis and example.
Fundamentally, a real symmetric matrix $\M$ is normal and thus diagonalizable, \textit{i.e.}, it can be eigen-decomposed into the following form:
\begin{align}
\M = \Q \bLambda \Q^{\top}
\label{eq:eigen}
\end{align}
where $\bLambda$ is a diagonal matrix with eigenvalues $\lambda_i$ along its diagonal, and $\Q$ contains the corresponding eigenvectors as columns.
Although the spectral theorem requires $\Q$ to be a unitary matrix, 
more generally, we can interpret \eqref{eq:eigen} to mean that symmetric real matrix $\M$ generalizes any diagonal matrix $\bLambda$ by pre- and post-multiplying it by any chosen square matrix $\Q$ and its transpose.
To demonstrate the importance of this generalization, consider the following simple example.
Define first the \textit{difference vector} as the difference between feature vectors $\f_i$ and $\f_j$, \ie, $\nabla \f_{i,j} = \f_i - \f_j$. 
Thus, given metric $\M$, the Mahalanobis distance between nodes $i$ and $j$ is computed as $(\f_i - \f_j)^{\top} \M (\f_i - \f_j) = \left( \nabla \f_{i,j} \right)^{\top} \M \left( \nabla \f_{i,j} \right)$.
Suppose now that there are only two available features $\f_i = [\f_i^{(1)}~ \f_i^{(2)}]^{\top}$ for every node $i$.
Suppose also that the optimal metric $\M$ in this case computes the difference of the two components in the difference vector $\nabla \f_{i,j}$, \ie,
\begin{equation}
\begin{split}
& (\nabla \f_{i,j})^{\top} \M (\nabla \f_{i,j}) \\
= & \left[ \begin{matrix}
\begin{smallmatrix}
\nabla \f_{i,j}^{(1)}~~ \nabla \f_{i,j}^{(2)} 
\end{smallmatrix}
\end{matrix} \right]
\underbrace{\left[ \begin{matrix}
\begin{smallmatrix}
1 & 1 \\
-1 & 1
\end{smallmatrix}
\end{matrix}
\right]}_{\Q}
\underbrace{\left[ \begin{matrix}
\begin{smallmatrix}
1 & 0 \\
0 & 0
\end{smallmatrix}
\end{matrix}
\right]}_{\bLambda}
\underbrace{\left[ \begin{matrix}
\begin{smallmatrix}
1 & -1 \\
1 & 1
\end{smallmatrix}
\end{matrix}
\right]}_{\Q^{\top}}
\left[ \begin{matrix}
\begin{smallmatrix}
\nabla \f_{i,j}^{(1)} \\
\nabla \f_{i,j}^{(2)}
\end{smallmatrix}
\end{matrix}
\right] \\ 
= & \left( \nabla \f_{i,j}^{(1)} - \nabla \f_{i,j}^{(2)} \right)^2
\label{eq:diffFeature}
\end{split}
\end{equation}

In this case, two nodes $i$ and $j$ with difference vector $\nabla \f_{i,j} = [\epsilon ~~ \epsilon]^{\top}$ having the same component $\nabla \f_{i,j}^{(1)} = \nabla \f_{i,j}^{(2)} = \epsilon$ will result in a Mahalanobis distance of $(\nabla \f_{i,j})^{\top} \M (\nabla \f_{i,j}) = 0$, no matter how large $\epsilon$ is.
On the other hand, \textit{any} non-zero PSD diagonal matrix $\bLambda' = \mathrm{diag}(\lambda_1', \lambda_2')$, where $\lambda_1' > 0$ or $\lambda_2' > 0$, will lead to a distance $(\nabla \f_{i,j})^{\top} \bLambda' (\nabla \f_{i,j}) = \lambda_1' \epsilon^2 + \lambda_2' \epsilon^2 = \infty$ as $\epsilon \rightarrow \infty$.
Thus, we can conclude that a diagonal-only metric $\bLambda'$ can be arbitrarily worse than the optimal metric $\M$ with off-diagonal terms, and off-diagonal terms for metric $\M$ are essential in computing feature distances.

To demonstrate that the above 2-feature example is not contrived, consider the following concrete application.
Suppose the first and second features,  $\f_i^{(1)}$ and $\f_i^{(2)}$, measure the $x$-location of a train on a line track at time $0$ and time $t > 0$, respectively.
Suppose the optimal metric $\M$ considers only the difference in \textit{velocity}, $\v_i - \v_j$, of the two trains $i$ and $j$, \ie,
\begin{align}
\left( \v_{i} - \v_j \right)^2 &= \left( (\f_i^{(1)} - \f_i^{(2)}) - (\f_j^{(1)} - \f_j^{(2)}) \right)^2 \\
&= \left( \nabla \f_{i,j}^{(1)} -  \nabla \f_{i,j}^{(2)} \right)^2
\end{align}
which is the same as \eqref{eq:diffFeature}.
Clearly, if the two trains $i$ and $j$ have the same velocity, the Mahalanobis distance between them should be zero, regardless of their difference in start / end locations. 
Using any diagonal-only metric, however, would compute a Mahalanobis distance that is a function of the difference between their start / end locations, which is incorrect.

\vspace{0.1in}
\subsubsection{Formulation}
We can now pose an optimization problem for $\mathbf{M}$ with GLR (\ref{eq:rgssp}) as objective: 
we seek the optimal metric $\mathbf{M}$ that yields the smallest GLR term given feature vector $\mathbf{f}_i$ for each node $i$, and one (or more) signal observation(s) $\mathbf{z}$.
Specifically, denote by $d_{i,j} = (z_i-z_j)^2$ the inter-node sample difference square of observation $\mathbf{z}$, we have 
\begin{equation}
\begin{split}
&\min_{\mathbf{M}}
    \sum_{\{i,j\}}\exp\left\{-(\mathbf{f}_i-\mathbf{f}_j)^{\top} \mathbf{M} (\mathbf{f}_i-\mathbf{f}_j) \right\} \, d_{i,j}\\
& \text{s.t.} \quad \, \mathbf{M} \succ 0.
\label{eq:optimize_L_new}
\end{split}
\end{equation}
If there are more than one signal observations $\mathbf{z}^1, \ldots, \mathbf{z}^S$ available, then $d_{i,j}$ in \eqref{eq:optimize_L_new} can be easily generalized to be the sum of inter-node sample difference squares of all observations, i.e. $d_{i,j} = \sum_{s=1}^S (z_i^s-z_j^s)^2$.

Minimizing (\ref{eq:optimize_L_new}) directly would lead to one pathological solution, i.e., $m_{i,i}=\infty, \forall i$, resulting in edge weights $w_{i,j} = 0$. 
Topologically, this means nodes in the graph are all isolated, defeating the goal of finding a similarity graph. 
To avoid this solution, we constrain the trace of $\mathbf{M}$ to be smaller than a constant parameter $C$, resulting in
\begin{equation}
\begin{split}
&\min_{\mathbf{M}}
    \sum_{\{i,j\}}\exp\left\{-(\mathbf{f}_i-\mathbf{f}_j)^{\top} \mathbf{M} (\mathbf{f}_i-\mathbf{f}_j) \right\} \, d_{i,j}\\
& \text{s.t.} \quad \,\mathbf{M} \succ 0; \;\;\;
\text{tr}(\mathbf{M}) \leq C.
\label{eq:optimize_c_constraint}
\end{split}
\end{equation}

One na\"{i}ve approach to the optimization problem in \eqref{eq:optimize_c_constraint} using proximal gradient descent \cite{Parikh31} is as follows.
The objective $q(\mathbf{M})$ in \eqref{eq:optimize_c_constraint} itself is convex and differentiable with respect to $\mathbf{M}$, and thus a gradient descent step $\nabla q(\mathbf{M})$ can be computed. 
The constraints in \eqref{eq:optimize_c_constraint} describe a feasible solution space that is a convex cone of all PD matrices with trace upper-bounded by $C$.
One can thus rewrite the constraints as a \textit{second} objective term $h(\mathbf{M})$ that evaluates to 0 if $\mathbf{M}$ is in the convex set and $\infty$ otherwise.
This convex but non-differentiable objective term $h(\mathbf{M})$ has the following proximal mapping $\mathrm{prox}_h(\mathbf{M})$:
orthogonally project the eigenvalues $\lambda_k$'s of $\mathbf{M}$ into the convex set: $\lambda_k > 0, \forall k$, $\sum_k \lambda_k \leq C$.
This results in a proximal gradient step:
$\mathbf{M}_{t+1} = \mathrm{prox}_h \left(\mathbf{M}_t - \gamma \nabla q(\mathbf{M}_t) \right)$, where $\gamma$ is a chosen step size.

However, this na\"{i}ve realization of proximal gradient requires eigen-decomposition of $\mathbf{M}_t$ per iteration $t$ with complexity $O(n^3)$, which is computation-expensive for large $\mathbf{M}_t$.
To completely circumvent eigen-decomposition, we rewrite the PD cone constraint as a set of linear constraints, which form another convex set (a polytope) that is much easier to solve.
In particular, our strategy is to optimize $\mathbf{M}$'s diagonal and off-diagonal entries alternately until convergence. 
We discuss the two optimizations in order next.

\subsection{Optimization of Diagonal Entries} 
\label{subsection:opt_diag}


When $\mathbf{M}$'s diagonal entries are optimized, \eqref{eq:optimize_c_constraint} can be simplified as follows.
Let $\mathbf{g}_{i,j} = \mathbf{f}_i-{\mathbf{f}}_j$, $\mathbf{g}_{i,j} \in \mathbb{R}^K$.
Further, let matrix $\mathbf{M}'$ be $\mathbf{M}$ with only the off-diagonal entries, i.e., $\mathbf{M}' = \mathbf{M} - \mathrm{diag}(\mathbf{M})$.
Then, 
\begin{equation}
\begin{split}
&\min_{\{m_{i,i}\}}
    \sum_{\{i,j\}}\exp\left\{-\mathbf{g}_{i,j}^{\top} \left( \mathbf{M}' + \mathrm{diag}(\mathbf{M}) \right) \mathbf{g}_{i,j} \right\} \, d_{i,j}\\
& \text{s.t.} \quad \,\mathbf{M} \succ 0; \;\;\;
\sum_{i} m_{i,i} \leq C.
\label{eq:optimize_diagonal}
\end{split}
\end{equation}

\subsubsection{Gershgorin-based reformulation}

To enforce the positive definiteness constraint $\mathbf{M} \succ 0$ using simple linear constraints, we leverage on the \textit{Gershgorin circle theorem} (GCT) \cite{gervsgorin1931uber} and constrain each Gershgorin disc $\Psi_i$ corresponding to each row $i$ of $\mathbf{M}$ to reside in strictly positive territory. 
Specifically, disc $\Psi_i$ has center $c_i = m_{i,i}$ and radius $r_i = \sum_{j\neq i} |m_{i,j}|$. 
To keep $\Psi_i$ in positive territory, we need the left-end $c_i - r_i$ to be positive. 
\eqref{eq:optimize_diagonal} thus becomes:
\begin{equation}
    \begin{split}
&\min_{\{m_{i,i}\}}
    \sum_{\{i,j\}}\exp \left\{-
    \sum_k m_{k,k} \mathbf{g}_{i,j}(k)^2 - 
    \mathbf{g}_{i,j}^{\top} \mathbf{M}' \mathbf{g}_{i,j} \right\} \, d_{i,j}\\
& \text{s.t.} \quad \,m_{i,i}-\sum_{j \neq i} |m_{i,j}| > 0, \forall i;
~~~~~~~\sum_{i} m_{i,i} \leq C.
\end{split}  
\end{equation}
We can further simplify the objective by defining $\tilde{d}_{i,j} = \exp \{ - \mathbf{g}_{i,j}^{\top} \mathbf{M}' \mathbf{g}_{i,j} \} d_{i,j}$, resulting in

\begin{equation}
    \begin{split}
&\min_{\{m_{i,i}\}}
    \sum_{\{i,j\}}\exp \left\{-
    \sum_k m_{k,k} \mathbf{g}_{i,j}(k)^2 \right\} \, \tilde{d}_{i,j}\\
& \text{s.t.} \quad \,m_{i,i}-\sum_{j \neq i} |m_{i,j}| > 0, \forall i;
~~~~~~~\sum_{i} m_{i,i} \leq C.
\label{eq:optimize_diagonal2}
\end{split}  
\end{equation}

\subsubsection{Proximal Gradient algorithm}

To solve \eqref{eq:optimize_diagonal2} efficiently, we employ a \textit{proximal gradient} (PG) approach \cite{Parikh31}. 
Let $\mathbf{m}=[m_{1,1},...,m_{K,K}]^{\top}$. The linear constraints for $\mathbf{m}$ form a convex set:
\begin{equation}
    \mathcal{S}=\left\{ \mathbf{m} ~\bigg|~ m_{i,i} > \sum_{j \neq i} |m_{i,j}|, \sum_{i=1}^K m_{i,i} \leq C, \forall i \right\}.
\end{equation}

Then, we define the indicator function $I_{\mathcal{S}}(\mathbf{m})$:
\begin{equation}
   I_{\mathcal{S}}(\mathbf{m})=
    \left\{
        \begin{array}{lr} 
            0, & \mathbf{m} \in \mathcal{S}  \\
            \infty, & \text{otherwise}
        \end{array}
    \right.
\end{equation}

We now rewrite the optimization for $\mathbf{m}$ as an unconstrained problem by exchanging the convex set constraint with indicator function $I_{\mathcal{S}}(\mathbf{m})$ in the objective: 
\begin{equation}
\min_{\mathbf{m}}
    \sum_{\{i,j\}}\exp \left\{-
    \sum_k m_{k,k} \mathbf{g}_{i,j}(k)^2 \right\} \, \tilde{d}_{i,j} + I_{\mathcal{S}}(\mathbf{m}).
    \label{eq:solve_m}
\end{equation}  
The first term is convex with respect to $\mathbf{m}$ and differentiable, while the second term $I_{\mathcal{S}}(\mathbf{m})$ is convex but non-differentiable. we can thus employ PG to solve (\ref{eq:solve_m}) as follows. 

We first compute the gradient of the first term with respect to $\mathbf{m}$:
\begin{equation}
\begin{split}
   & \bigtriangledown F(\mathbf{m})= \\
   & \left[
\begin{array}{c}
   -\sum\limits_{\{i,j\}} \exp\left\{-
    \sum\limits_k m_{k,k} \mathbf{g}_{i,j}(k)^2 \right\} \mathbf{g}_{i,j}(1)^{2} \tilde{d}_{i,j} \\
   \vdots \\
   -\sum\limits_{\{i,j\}} \exp\left\{-
    \sum\limits_k m_{k,k} \mathbf{g}_{i,j}(k)^2 \right\} \mathbf{g}_{i,j}(K)^{2} \tilde{d}_{i,j}
  \end{array}
  \right]
\end{split}
\end{equation}  
We next define a proximal mapping $\Pi_{I_{\mathcal{S}}}(\mathbf{v})$ for the second term---indicator function $I_{\mathcal{S}}(\mathbf{v})$---which is a projection onto the convex set $\mathcal{S}$, i.e.,
\begin{equation}
    \Pi_{I_{\mathcal{S}}}(\mathbf{v})=
    \left\{
        \begin{array}{lr} 
            P_{\mathcal{T}}(\mathbf{v}), & \vec{1}^{\top} P_{\mathcal{T}}(\mathbf{v}) \leq C,  \\
            P_{\mathcal{T}}(\mathbf{v}-\alpha\cdot\vec{1}), & \text{otherwise,}
        \end{array}
    \right.
\end{equation}
where $P_{\mathcal{T}}(\mathbf{v})=\max\left\{ v_i, \sum_{j \neq i}|m_{i,j}| \right\} _{i=1}^{K}$, and $\alpha$ is any positive root of $\vec{1}^{\top}P_{\mathcal{T}}(\mathbf{v}-\alpha\cdot\vec{1}) = C$ \cite{beck2017first}.

Each iteration of the PG algorithm can be now written as:
\begin{equation}
    \mathbf{m}^{l+1}=\Pi_{I_{\mathcal{S}}}(\mathbf{m}^{l}-t\bigtriangledown F(\mathbf{m}^{l})),
\end{equation}
where $t$ is the step size. As discussed in \cite{Parikh31}, 
the algorithm will converge with rate $O(1/l)$ for a fixed step size $t^l = t \in (0,2/L]$, where $L$ is a Lipschitz constant that requires computation of the Hessian of $F$. 
In our experiment, we choose a small step size $t$ empirically, which is small enough to satisfy the Lipschitz smoothness of the objective function.
In the first iteration, we initialize $\mathbf{M}$ to be a diagonal matrix with each diagonal entry $m_{i,i}=C/K$, thus ensuring $\mathbf{M}$ is PD and $\mathrm{tr}(\mathbf{M}) \leq C$. 

Further, we may reduce the complexity of the proposed algorithm via \textit{accelerated proximal gradient} (APG) \cite{beck09tip,li15apg}. APG is able to accelerate the convergence, by first extrapolating a point from the current point and the previous point and then performing a proximal gradient step. We leave this as our future work.

\subsection{Optimization of Off-diagonal Entries}
\label{subsec:off_diag_entries}


For off-diagonal entries of $\mathbf{M}$, we develop a block coordinate descent algorithm, which optimizes one row / column at a time. 

\vspace{0.1in}
\subsubsection{Block Coordinate Iteration}
 
First, we divide $\mathbf{M}$ into four sub-matrices:
\begin{equation}
\mathbf{M} = \begin{bmatrix}
m_{1,1} & \mathbf{M}_{1,2} \\
\mathbf{M}_{2,1} & \mathbf{M}_{2,2} 
\end{bmatrix},
\label{eq:submatrix}
\end{equation}
where $m_{1,1} \in \mathbb{R}$, $\mathbf{M}_{1,2} \in \mathbb{R}^{1 \times (K-1)}$, $\mathbf{M}_{2,1} \in \mathbb{R}^{(K-1) \times 1}$ and $\mathbf{M}_{2,2} \in \mathbb{R}^{(K-1) \times (K-1)}$. 
The assumption that $\mathbf{M}$ is symmetric means $\mathbf{M}_{1,2} = \mathbf{M}_{2,1}^{\top}$.     
Our strategy is to optimize one row and column of off-diagonal entries represented by $\mathbf{M}_{2,1}$ in one iteration, given objective and constraints in \eqref{eq:optimize_c_constraint}. 
In the next iteration, a different row and column is selected, and with appropriate rows and columns reordering, the optimization variable $\mathbf{M}_{2,1}$ can still reside in the first row and column as shown in \eqref{eq:submatrix}.

By the \textit{Haynsworth inertia additivity} \cite{haynsworth1968determination}, a symmetric real matrix $\mathbf{M}$ is PD if and only if both its sub-matrix $\mathbf{M}_{2,2}$ and its corresponding Schur complement $m_{1,1} - \mathbf{M}_{2,1}^{\top} \mathbf{M}_{2,2}^{-1} \mathbf{M}_{2,1}$ are PD. 
Hence, we can ensure $\mathbf{M}$ is PD by constraining the Schur complement  $m_{1,1} - \mathbf{M}_{2,1}^{\top} \mathbf{M}_{2,2}^{-1} \mathbf{M}_{2,1}$ to be positive, given that matrix $\mathbf{M}$, and therefore sub-matrix $\mathbf{M}_{2,2}$, are both PD from the previous iteration.

In the first iteration, we initialize $\mathbf{M}$ to be a diagonal matrix with diagonal entries as optimized in Sec.~\ref{subsection:opt_diag}.
In each subsequent iteration, we impose a positivity constraint on the Schur complement of a submatrix $\mathbf{M}_{2,2}$ as follows:
\begin{equation}
    m_{1,1} - \mathbf{M}_{2,1}^{\top} \mathbf{M}_{2,2}^{-1} \mathbf{M}_{2,1} > 0.  
\end{equation}
Optimization problem \eqref{eq:optimize_c_constraint} thus becomes:   
\begin{equation}
\begin{split}
&\min_{\mathbf{M}_{2,1}}
    \sum_{\{i,j\}}\exp\left\{-({\mathbf{f}}_i-{\mathbf{f}}_j)^{\top} \mathbf{M}({\mathbf{f}}_i-{\mathbf{f}}_j) \right\}d_{i,j}\\
& \text{s.t.} \quad \,m_{1,1}-\mathbf{M}_{2,1}^{\top} \mathbf{M}_{2,2}^{-1} \mathbf{M}_{2,1} > 0, \\ 
& ~~~~~~~m_{1,1} \leq C - \text{tr}(\mathbf{M}_{2,2}).
\end{split}  
\label{eq:schur}
\end{equation}

Given $\mathbf{M}_{2,2}$ is fixed in \eqref{eq:schur},  we can simplify the objective as follows. 
Writing $\mathbf{M}$ in terms of its four sub-matrices in \eqref{eq:submatrix}, we simplify the objective via matrix multiplication as 
\begin{equation}
    \sum_{\{i,j\}} 
\exp\left\{-2 \mathbf{g}_{i,j}(1) \mathbf{M}_{2,1}^{\top} \mathbf{g}_{i,j}(2))\right\}\tilde{d}_{i,j}
\end{equation}
where the reused notation $\tilde{d}_{i,j}=\exp\{-\mathbf{g}_{i,j}(1)^2 m_{1,1}-\mathbf{g}_{i,j}(2)^{\top} \mathbf{M}_{2,2} \mathbf{g}_{i,j}(2) \}d_{i,j}$ is a constant as $m_{1,1}$ and $\mathbf{M}_{2,2}$ are fixed in the iteration.
$\mathbf{g}_{i,j}(1)$ denotes the first entry in vector $\mathbf{g}_{i,j}$, and $\mathbf{g}_{i,j}(2) \in \mathbb{R}^{K-1}$ denotes the remaining entries.

\vspace{0.1in}
\subsubsection{$\lambda_{\max}$-bounded Reformulation}

Computation of a large matrix inverse $\mathbf{M}_{2,2}^{-1}$ in \eqref{eq:schur} per iteration is costly.
To avoid computing $\mathbf{M}_{2,2}^{-1}$, we derive a bound based on the largest eigenvalue $\lambda_{\max}$ of $\mathbf{M}_{2,2}^{-1}$ to ensure the positivity constraint on the Schur complement $m_{1,1} - \mathbf{M}_{2,1}^{\top} \mathbf{M}_{2,2}^{-1} \mathbf{M}_{2,1}$ in \eqref{eq:schur} is satisfied. 

First, since $\mathbf{M}_{2,2}^{-1}$ is a real and  symmetric PD matrix, it admits diagonalization with eigen-matrix $\mathbf{U}$ (eigenvectors as columns) and diagonal matrix $\boldsymbol{\Lambda}$ with real eigenvalues $0 < \lambda_{1} \leq \ldots \leq \lambda_{K}$ along its diagonal. 
Hence,
\begin{equation}
    \begin{split}
    \mathbf{M}_{2,1}^{\top} \mathbf{M}_{2,2}^{-1} \mathbf{M}_{2,1} 
    & = \mathbf{M}_{2,1}^{\top} \mathbf{U} \boldsymbol{\Lambda} \mathbf{U}^{\top} \mathbf{M}_{2,1} \\
    & = (\mathbf{U}^{\top} \mathbf{M}_{2,1})^{\top} \boldsymbol{\Lambda} (\mathbf{U}^{\top} \mathbf{M}_{2,1}),
    \end{split}
\end{equation}
which is essentially scaling the $l_2$-norm of $\mathbf{U}^{\top} \mathbf{M}_{2,1}$ by eigenvalues in $\boldsymbol{\Lambda}$. 
Hence, a sufficient condition to the first constraint in (\ref{eq:schur}) is to bound with the maximum eigenvalue $\lambda_{\text{max}}$ of $\mathbf{M}_{2,2}^{-1}$: 
\begin{equation}
\begin{split}
    & m_{1,1} > \lambda_{\text{max}} (\mathbf{U}^{\top} \mathbf{M}_{2,1})^{\top} (\mathbf{U}^{\top} \mathbf{M}_{2,1}) \\
    & \Rightarrow m_{1,1} > \lambda_{\text{max}} \mathbf{M}_{2,1}^{\top} \mathbf{U} \mathbf{U}^{\top} \mathbf{M}_{2,1} \\
    & \Rightarrow m_{1,1} > \lambda_{\text{max}} \mathbf{M}_{2,1}^{\top} \mathbf{M}_{2,1},
\end{split}
\end{equation}

$\lambda_{\text{max}}$ is the reciprocal of the minimum eigenvalue $\theta_{\text{min}}$ of $\mathbf{M}_{2,2}$, i.e., $\lambda_{\text{max}}=\frac{1}{\theta_{\text{min}}}$. 
We employ Locally Optimal Block Preconditioned Conjugate Gradient (LOBPCG) \cite{Knyazev01} to calculate $\theta_{\text{min}}$, which is efficient to compute the extreme eigen-pairs of a large sparse matrix with linear convergence.


Having computed $\theta_{\min}$ from $\mathbf{M}_{2,2}$ without eigen-decomposition, we can reformulate (\ref{eq:schur}) as    
\begin{equation}
\begin{split}
&\min_{\mathbf{M}_{2,1}}
    \sum_{\{i,j\}} 
\exp\left\{- 2 \mathbf{g}_{i,j}(1) \mathbf{M}_{2,1}^{\top} \mathbf{g}_{i,j}(2)\right\}\tilde{d}_{i,j}\\
& \text{s.t.} \quad \,m_{1,1} > \frac{1}{\theta_{\min}} \mathbf{M}_{2,1}^{\top} \mathbf{M}_{2,1}.
\end{split}  
\label{eq:constrained_final}
\end{equation}
Next, we design an efficient algorithm to address (\ref{eq:constrained_final}). 

\vspace{0.1in}
\subsubsection{PG Algorithm}

In each iteration of the block descent algorithm, we hold the diagonal entries fixed and optimize $\mathbf{M}_{2,1}$ in (\ref{eq:constrained_final}).  

The objective is convex while the lone constraint for $\mathbf{M}_{2,1}$ forms a convex set. 
In particular, the constraint in (\ref{eq:constrained_final}) reduces to 
\begin{equation}
    \mathcal{S}=\{\| \mathbf{M}_{2,1} \|_2^2 < \theta_{\text{min}} m_{1,1}\},
\end{equation}
which is a $(K-1)$-dimensional norm ball with radius $\sqrt{\theta_{\text{min}} m_{1,1}}$. 
We now define an indicator function $I_{\mathcal{S}'}(\mathbf{M}_{2,1})$:
\begin{equation}
   I_{\mathcal{S}'}(\mathbf{M}_{2,1})=
    \left\{
        \begin{array}{lr} 
            0, & \mathbf{M}_{2,1} \in \mathcal{S}'  \\
            \infty, & \text{otherwise}
        \end{array}
    \right. 
\end{equation}
We can rewrite the optimization for $\mathbf{M}_{2,1}$ as an unconstrained problem by exchanging the convex set constraint with the indicator function $I_{\mathcal{S}'}(\mathbf{M}_{2,1})$  in the objective:
\begin{equation}
\min_{\mathbf{M}_{2,1}}
    \sum_{\{i,j\}} 
    \exp\left\{- 2 \mathbf{g}_{i,j}(1) \mathbf{M}_{2,1}^{\top} \mathbf{g}_{i,j}(2)\right\}\tilde{d}_{i,j} 
    + I_{\mathcal{S}'}(\mathbf{M}_{2,1}).
    \label{eq:solve_M21}
\end{equation}  
The first term is convex with respect to $\mathbf{M}_{2,1}$ and differentiable, while the second term $I_{\mathcal{S}'}(\mathbf{M}_{2,1})$ is convex but non-differentiable. 
Hence, we employ again the PG algorithm to solve (\ref{eq:solve_M21}).   

Specifically, we first compute the gradient of the first term $F$ in the objective with respect to $\mathbf{M}_{2,1}$ as 
\begin{equation}
\begin{split}
    & \bigtriangledown F(\mathbf{M}_{2,1}) = \\
    & -2 \sum_{\{i,j\}} \mathbf{g}_{i,j}(1) \mathbf{g}_{i,j}(2)\exp\left\{- 
      2 \mathbf{g}_{i,j}(1) \mathbf{M}_{2,1}^{\top} \mathbf{g}_{i,j}(2)\right\}\tilde{d}_{i,j},
\end{split}
\end{equation}
which will be adopted in the step of gradient descent. 
We define a proximal mapping $\Pi_{I_{\mathcal{S}'}}(\mathbf{v})$ for $\mathbf{v} \in \mathbb{R}^{(K-1) \times 1} $, which is a projection onto the norm ball with radius $\sqrt{\theta_{\text{min}} m_{1,1}}$:
\begin{equation}
     \Pi_{I_{\mathcal{S}'}}(\mathbf{v}) =
    \left\{
        \begin{array}{lr} 
            \mathbf{v}, & \| \mathbf{v} \|_2 \leq \sqrt{\theta_{\text{min}} m_{1,1}}  \\
            \frac{\mathbf{v}}{\| \mathbf{v} \|_2} \cdot \sqrt{\theta_{\text{min}} m_{1,1}}, & \text{otherwise.}
        \end{array}
    \right. 
\end{equation}
Then each iteration in the PG algorithm can be written as
\begin{equation}
     \mathbf{M}_{2,1}^{l+1} := \Pi_{I_{\mathcal{S}'}}(\mathbf{M}_{2,1}^l - t \bigtriangledown F(\mathbf{M}_{2,1}^l)),
\end{equation}
where $t$ is the step size as discussed in Sec.~\ref{subsection:opt_diag}. 
We may also deploy APG to reduce the complexity further.

Finally, we analyze the convergence of our algorithm. 
The proposed alternating optimization algorithm optimizes diagonal and off-diagonal terms in $\mathbf M$ in turn. 
When computing a solution for diagonal or off-diagonal terms, we adopt the newly computed solution only if the objective \textit{strictly} decreases. 
Further, using an exponential kernel to compute edge weights means that any metric $\mathbf{M}$ would always result in non-negative edge weights. 
This results in a PSD graph Laplacian $\mathbf{L}$ \cite{cheung2018graph}, and our GLR objective \eqref{eq:optimize_c_constraint}
is lower-bounded by $0$. 
Hence, our algorithm strictly decrements an objective iteratively that is lower-bounded by $0$, and 
our algorithm converges to a locally optimal solution.


\section{Feature Metric Learning for Point Cloud Denoising}
\label{sec:formulation}
Having described our feature graph learning scheme, we now employ it for 3D point cloud denoising. 
We first propose a patch-based model and designate graph connectivities over similar patches in a neighborhood. 
Then we formulate an inverse problem for point cloud denoising with GLR for regularization, which can be interpreted alternatively assuming an IGMRF model.
Finally, we develop an alternating algorithm to efficiently solve the formulated problem.

\subsection{Patch-based Model}

We assume an additive noise model for a point cloud, namely
\begin{equation}
    \vec{P} = \vec{Y} + \vec{E},
    \label{eq:model1}
\end{equation}
where $\vec{P} \in \mathbb{R}^{N \times 3} $ denotes the observed noise-corrupted 3D coordinates of the target point cloud with $N$ points, $\vec{Y} \in \mathbb{R}^{N \times 3}$ is the ground truth 3D coordinates of the point cloud, and $\vec{E} \in \mathbb{R}^{N \times 3}$ is an additive noise. 

Modeling of $\mathbf{E}$ depends on the actual point cloud acquisition mechanism. 
There exist a wide range of point cloud acquisition systems at different price points---from consumer-level depth sensors like Intel RealSense\footnote{https://www.intelrealsense.com/} costing 150 USD to high-end outdoor scanners like Teledyne Optech\footnote{https://www.teledyneoptech.com/en/products/static-3d-survey/} that cost up to 250,000 USD---and defining accurate noise models for all of them is difficult. 
We thus select the most common Gaussian noise model, which has been shown to be reasonably accurate for popular depth cameras like Microsoft Kinect \cite{nguyen12, xiafang08}.
Hence, $\vec{E}$ in (\ref{eq:model1}) represents zero-mean \textit{additive white Gaussian noise} (AWGN) with standard deviation $\sigma$, i.e.,
\begin{equation}
    \vec{E} \sim \mathcal{N}(\mathbf{0},\sigma^2 \mathbf{I}).
    \label{eq:noise_model}
\end{equation}

In image denoising, a simple and effective assumption is \textit{self-similarity} (also known as \textit{nonlocal means} (NLM)) \cite{buades2005non, dabov2007image, rosma13}: similar pixel patches exist throughout the same image, which can be searched and gathered for joint denoising.
To exploit self-similarity also in point clouds, we also divide $\vec{P}$ into \textit{patches}.
However, defining inter-patch similarity in a point cloud is not straightforward, because a point cloud is a collection of irregularly sampled points in 3D space\footnote{\cite{zeng18arxiv} formally defined patch similarity based on projections on parallel planes, but it is computation-intensive.}. 
Instead, our work circumvents explicit search for similar patches.

Mimicking a fixed-size image patch with a center pixel, we first define a patch $\vec{V}_i$ in a point cloud $\vec{P}$ as done in \cite{zeng18arxiv}:

\vspace{0.1in}
\noindent
\textbf{Definition 1.} A patch $\vec{V}_i$ in a point cloud is a local point set of $k+1$ points,
consisting of a center point $\vec{c}_i \in \mathbb{R}^3$ and its $k$-nearest-neighbors in terms of Euclidean distance. 


\vspace{0.1in}
We divide the input point cloud $\vec{P}$ into a set of $M$ \textit{overlapping} patches, each with a center point $\vec{c}_i \in \vec{P}$. 
The patch centers are selected from a subset of points in $\mathbf{P}$. 
Intuitively, uniformly distributed center points are preferred, since they efficiently cover the entire point cloud. 
Hence, we employ the uniform sampling method in \cite{eldar97}.
$M$ is empirically set, where $M \leq N$ and $(k+1)M \geq N$.   

Next, for each center point $\mathbf{c}_i$ we construct a patch by identifying $\mathbf{c}_i$'s $k$ nearest neighbors. 
To exploit the assumed similarity among patches for denoising, we align patches via translation so each patch has its center at the origin.
This results in patch set $\vec{V} \in \mathbb{R}^{(k+1)M \times 3}$:
\begin{equation}
    \vec{V} = \mathcal{T} \vec{Y} - \vec{C},
    \label{eq:patch}
\end{equation}
where $\mathcal{T} \in \{0, 1\}^{(k+1)M \times N}$ is a selection matrix to choose points from $\vec{Y}$ to form $M$ patches, each with $k+1$ points. 
Specifically, each row in $\mathcal{T}$ contains only 0s except one 1 to choose one point in $\vec{Y}$.  
$\vec{C}=\{\vec{c}_i\}_{i=1}^M \in \mathbb{R}^{(k+1)M \times 3}$ denotes the coordinates of patch centers.
 



\subsection{Proposed Graph Connectivity}
\label{subsec:graph_construction}

We connect two adjacent patches as follows. 
Two patches are considered adjacent if their centers are $k$-nearest neighbors.  
Specifically, we use the $k$-nearest-neighbor ($k$NN) algorithm to search the nearest $\varepsilon$ patches of each patch as the neighbors based on the Euclidean distance between patch centers. 
Then we build a graph over each pair of adjacent patches.  
Overall, this leads to a $\varepsilon$-nearest-patch graph on the entire point cloud. 

\begin{figure}
    \centering
    \includegraphics[width=0.3\textwidth]{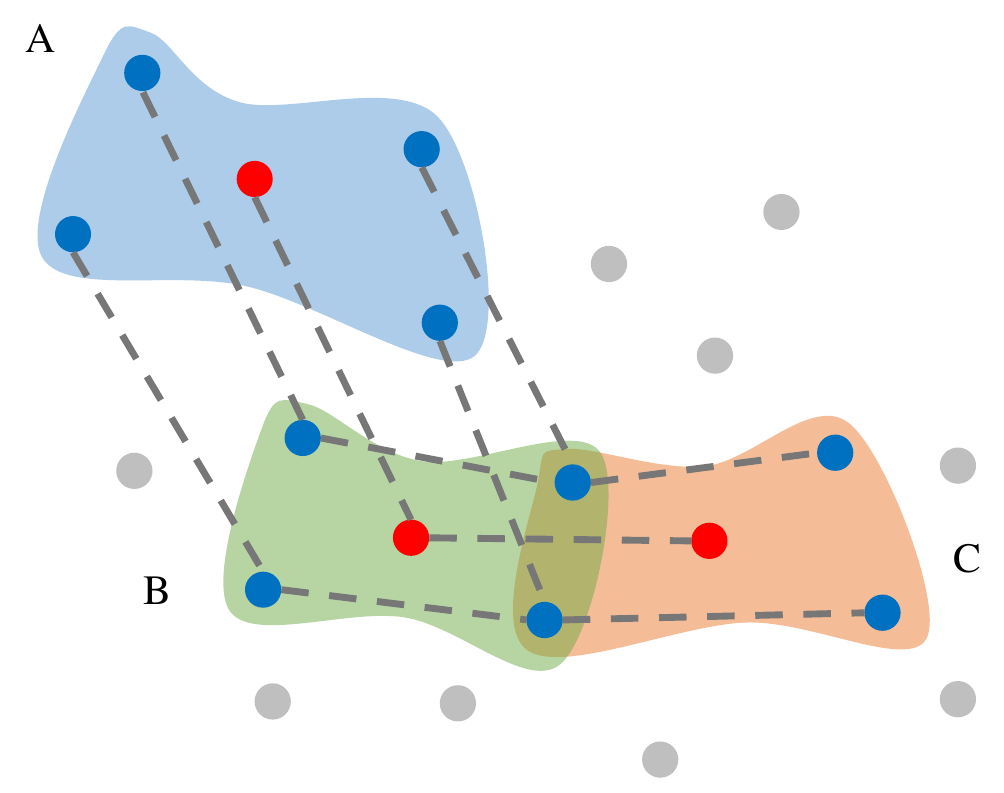}
    \caption{Illustration of graph connectivities over adjacent patches. Corresponding points in each pair of neighboring patches are connected, while there are no connectivities within each patch unless when patches are overlapped.}
    \label{fig:patch_graph}
\end{figure}

Each point in one patch is then connected to its \textit{corresponding point} in the other patch.
For simplicity, we treat a pair of points in adjacent patches as corresponding points if their coordinates \textit{relative} to their respective centers are closest to each other. 
Namely, for each point $\mathbf{p}_i \in \mathbf{V}_s$, we search the corresponding nearest point $\mathbf{p}_j \in \mathbf{V}_t$, which has the smallest Euclidean distance to $\mathbf{p}_i$:
\begin{equation}
    \mathbf{p}_j = \argmin_{\mathbf{p}_l \in \mathbf{V}_t} \|\mathbf{p}_i-\mathbf{p}_l\|_2,~~ \mathbf{p}_i \in \mathbf{V}_s.
\end{equation}
We do not explicitly connect points within the same patch, though two points in a patch may nonetheless be connected because they are also corresponding points in two different patches due to patch overlaps, as shown in Fig.~\ref{fig:patch_graph}. 

We note that this is one possible graph connectivity among many. 
For example, one can in addition enable intra-patch filtering by drawing connections among points in the same patch \cite{dinesh18arxiv,duan2018weighted}, resulting in a denser graph. 
For simplicity, we employ the most basic graph connectivity given inter-patch similarities. 
More sophisticated graph constructions considering also intra-patch filtering is left for future work.

Next, we formulate the problem of point cloud denoising via MAP estimation based on the chosen graph connectivities.

\subsection{MAP Formulation for Point Cloud Denoising}

We pose a MAP estimation problem for the underlying patches $\vec{V}$: given the observed patches $\hat{\vec{V}}$, find the most probable signal $\vec{V}$,
\begin{equation}
    \vec{\tilde{\mathcal{V}}}_\text{MAP}(\hat{\vec{V}}) = \argmax_{\vec{V}} \ f(\hat{\vec{V}} \mid \vec{V})g(\vec{V}),
    \label{eq:final_map}
\end{equation}
where $f(\hat{\vec{V}} \mid \vec{V})$ is the likelihood function, and $g(\vec{V})$ is the prior probability distribution of $\vec{V}$. 

\subsubsection{Likelihood function}

Since patches are extracted from the observed point cloud, $f(\hat{\vec{V}} \mid \vec{V})$ is equivalent to $f(\mathbf{P} \mid \mathbf{Y})$. We thus define the likelihood function according to the additive Gaussian noise model in (\ref{eq:model1}) and (\ref{eq:noise_model}):
\begin{equation}
    f(\hat{\vec{V}} \mid \vec{V})=f(\mathbf{P} \mid \mathbf{Y})=\exp \left\{-\alpha \left\| \mathbf{Y}-\mathbf{P} \right\|_F^2\right\},
\label{eq:likelihood}    
\end{equation}
where $\alpha=1/(2\sigma^2)$ is a parameter. 

\subsubsection{Prior probability distribution}
\label{subsec:prior}

We employ signal-dependent GLR as the prior, namely, 
\begin{equation}
    g(\vec{V})=  \text{exp}\left\{-\beta\mathrm{tr}\left(\vec{V}^{\top}\mathbf{L}(\mathbf{M}) \vec{V}\right)\right\},
\label{eq:prior_L}
\end{equation}
where $\beta$ is a parameter, and $\mathbf{M}$ is the Mahalanobis distance matrix satisfying constraints in (\ref{eq:optimize_c_constraint}). 
This prior essentially enforces the smoothness of patch signals $\vec{V}$ with respect to the underlying graph $\mathbf{L}$.  

We provide an alternative interpretation of GLR next.
Specifically, we model the self-similarity among chosen patches in point clouds via \textit{first-order IGMRFs on irregular lattices}.
The definition of IGMRF is as follows \cite{rue2005gaussian}: 

\vspace{0.1in}
\noindent
\textbf{Definition 2.} Let $\mathbf{Q}$ be an $n \times n$ symmetric positive \textit{semi-definite} matrix with rank $n-r>0$. Then $\vec{x}=(x_1,...,x_n)^{\top}$ is an intrinsic GMRF of order $r \geq 0$ with parameters $(\vec{\mu},\mathbf{Q})$, if its density is 
\begin{equation}
    \pi(\vec{x}) = (2\pi)^{-\frac{n-k}{2}}(|\mathbf{Q}|^*)^{\frac{1}{2}} \exp\left\{-\frac{1}{2}(\vec{x}-\vec{\mu})^{\top}\mathbf{Q}(\vec{x}-\vec{\mu})\right\},
    \label{eq:igmrf}
\end{equation}
where $|\cdot|^*$ denotes the \textit{generalized determinant} (the product of non-zero eigenvalues). 
An intrinsic GMRF of order $r \geq 0$ is also known as an improper GMRF of rank $n-r$.   

Further, $\vec{x}$ is an intrinsic GMRF with respect to a graph $\mathcal{G}=\{\mathcal{V},\mathcal{E}\}$, where
\begin{equation}
    Q_{i,j} \neq 0 \Longleftrightarrow \{i,j\} \in \mathcal{E},~\forall i \neq j.
\end{equation}

In our context, we only consider the case where the graph $\mathcal{G}$ is \textit{loopless} and \textit{connected}, leading to first-order IGMRF modelling of point clouds (i.e., $r=1$) \cite{chung97}. 

Specifically, for patches translated to the origin $\vec{V} = \mathcal{T} \vec{Y} - \vec{C}$, we model the difference between corresponding points $\vec{p}_i$ and $\vec{p}_j$ in adjacent patches. 
As the coordinates along each axis are independent, we consider each component separately. 
Taking the $x$-axis coordinate $x_i$ and $x_j$ for instance, we define a normal increment 
\begin{equation}
    x_i - x_j \sim \mathcal{N}(0,1/(w_{i,j}\kappa)),
\end{equation}
where $\kappa$ is a precision parameter, and $ w_{i,j}$ is a positive and symmetric weight we incorporate for each pair of neighboring nodes $i$ and $j$. The joint density then becomes
\begin{equation}
    \pi(\vec{x}) \propto \kappa^{(n-1)/2}\exp\left\{-\frac{\kappa}{2}\sum_{i \sim j} w_{i,j}(x_i-x_j)^2\right\}.
\end{equation}

As derived in \cite{rue2005gaussian}, the corresponding precision matrix has the following form:
\begin{equation}
    Q_{i,j}=\kappa
    \left\{
        \begin{array}{ll} 
            \sum_{k \sim i} w_{i,k} & i=j,  \\
            -w_{i,j} & i \sim j, \\
            0 & \text{otherwise.}
        \end{array}
    \right.
    \label{eq:weighted_q}
\end{equation}

Hence, under the first-order IGMRF model on irregular lattices where we assume normal increments between corresponding points, the prior distribution of $\vec{V}$ is 
\begin{equation}
\begin{split}
    g(\vec{V}) & =\beta  \exp\left\{-\frac{1}{2}\vec{V}^{\top}\mathbf{Q} \vec{V}\right\} \\
    & \propto \kappa^{(n-1)/2}\exp\left\{-\frac{\kappa}{2}\sum_{i \sim j} w_{i,j}(x_i-x_j)^2\right\},
    \label{eq:prob_func}
\end{split}
\end{equation}
where $\beta=(2\pi)^{-\frac{n-1}{2}}(|\mathbf{Q}|^*)^{\frac{1}{2}}$. Here $\mathbf{Q}$ has the specific form as in (\ref{eq:weighted_q}).

Further, comparing the specific form of $\vec{Q}$ in (\ref{eq:weighted_q}) and the definition of the combinatorial graph Laplacian $\mathbf{L}$ in Sec.~\ref{subsec:laplacian}, we have 
\begin{equation}
    \mathbf{Q} = \kappa \mathbf{L}.
\end{equation}

Hence, we replace $\mathbf{Q}$ in the quadratic term in (\ref{eq:prob_func}) with $\mathbf{L}$ and consider all the three components, which leads to the GLR prior in (\ref{eq:prior_L}) and thus an alternative perspective of GLR under first-order IGMRF. 


\subsubsection{Final formulation for point cloud denoising}

Combining (\ref{eq:patch}), (\ref{eq:final_map}), (\ref{eq:likelihood}) and (\ref{eq:prior_L}), we have
\begin{equation}
\begin{aligned}
    \max_{\vec{Y},\mathbf{M}} \exp \left\{ \right. & -\alpha \left\| \mathbf{Y}-\mathbf{P} \right\|_F^2 \\
    & \left. -\beta \mathrm{tr}\left((\mathcal{T} \vec{Y} - \mathbf{C})^{\top}\mathbf{L}(\mathbf{M})(\mathcal{T} \vec{Y} - \mathbf{C})\right) \right\}.
\end{aligned}
\label{eq:formulation}
\end{equation}
We rewrite the objective by taking the logarithm of (\ref{eq:formulation}) and multiplying by $-1$. Also, with the constraints of $\mathbf{M}$ considered, we have the final problem formulation for point cloud denoising
\begin{equation}
\begin{split}
&\min_{\vec{Y},\mathbf{M}}
    \| \vec{Y} -  \vec{P} \|_F^2 + \gamma  \mathrm{tr}\left((\mathcal{T} \vec{Y} - \mathbf{C})^{\top}\mathbf{L}(\mathbf{M}) (\mathcal{T} \vec{Y} - \mathbf{C})\right), \\
& \text{s.t.} \quad \,\mathbf{M} \succ 0; \;\;\;
\mathrm{tr}(\mathbf{M}) \leq C,
\end{split}
\label{eq:formulation_final}
\end{equation}
where $\gamma= \beta / \alpha$. 

Next, we develop an alternating algorithm to solve (\ref{eq:formulation_final}).


\subsection{Proposed Algorithm for Point Cloud Denoising}
\begin{figure*}
    \centering
    \includegraphics[width=1.0\textwidth]{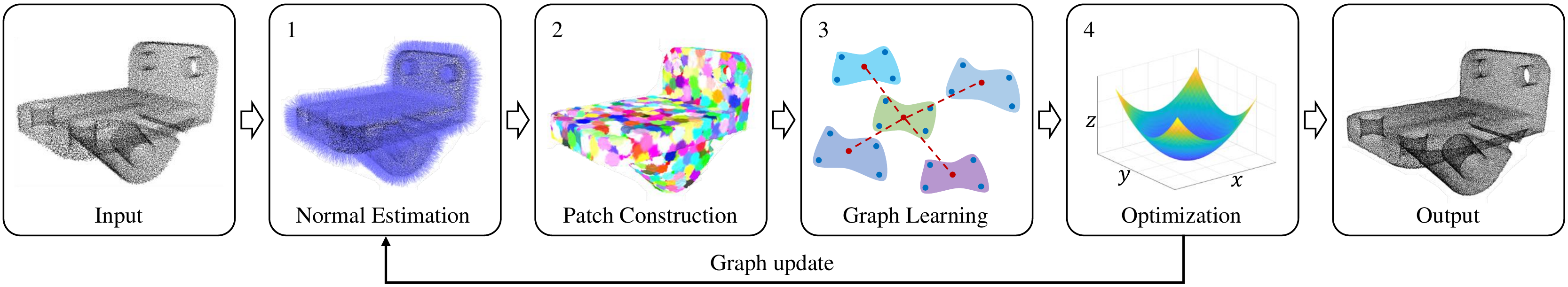}
    \caption{The flowchart of the proposed point cloud denoising algorithm. }
    \vspace{-0.1in}
    \label{fig:flowchart}
\end{figure*}

We propose to address (\ref{eq:formulation_final}) by alternately optimizing the point cloud $\mathbf{Y}$ and the Mahalanobis distance matrix $\mathbf{M}$. 
The iterations terminate when the difference in the objective between two consecutive iterations stops decreasing. 
Parameter settings are presented in Section~\ref{sec:results}.

\subsubsection{Optimizing the point cloud $\mathbf{Y}$}

In the first iteration, we initialize $\mathbf{M}$ with an identity matrix and thus fix $\mathbf{L}$ in (\ref{eq:formulation_final}). 
Next, taking the derivative of (\ref{eq:formulation_final}) with respect to the three components of $\vec{Y}$, $\{\vec{Y}_x,\vec{Y}_y,\vec{Y}_z\}$, we have
\begin{equation}
\begin{array}{c}
    (\gamma \mathcal{T}^{\top}\mathbf{L}\mathcal{T} + \mathbf{I})\vec{Y}_x=\vec{P}_x+\gamma \mathcal{T}^{\top} \mathbf{L} \mathbf{C}_x, \\
    (\gamma \mathcal{T}^{\top}\mathbf{L}\mathcal{T} + \mathbf{I})\vec{Y}_y=\vec{P}_y+\gamma \mathcal{T}^{\top} \mathbf{L} \mathbf{C}_y, \\
    (\gamma \mathcal{T}^{\top}\mathbf{L}\mathcal{T} + \mathbf{I})\vec{Y}_z=\vec{P}_z+\gamma \mathcal{T}^{\top} \mathbf{L} \mathbf{C}_z,
\end{array}
\label{eq:sol}
\end{equation}
where $\mathbf{I}$ is an identity matrix. 
(\ref{eq:sol}) can be treated as three linear equation sets and thus can be efficiently solved using conjugate gradient (CG) methods, such as the LSQR algorithm \cite{paige1982lsqr}. 
The acquired solution of $\vec{Y}$ is then used to update $\mathbf{M}$ in the subsequent iteration. 

\subsubsection{Optimizing the Mahalanobis distance matrix $\mathbf{M}$}

When $\vec{Y}$ is fixed, the optimization of $\mathbf{M}$ is feature metric learning in Sec.~\ref{sec:learning}. $\mathbf{L}$ is then computed from optimized $\mathbf{M}$ by definition and our edge weight kernel.


Specifically, we consider two features: Cartesian coordinates and normals. 
As done in \cite{dinesh18arxiv, duan2018weighted}, surface normals are adopted to promote smoothness of the underlying surface on which the point clouds are discrete samples. 
Along with coordinates, we thus form a $6$-dimensional feature vector at each point $i$, i.e., $\vec{f}_i = [x_i,y_i,z_i,n_x^i,n_y^i,n_z^i]^{\top}$, where $[x_i,y_i,z_i]$ denotes the coordinates of point $i$, and $[n_x^i,n_y^i,n_z^i]$ denotes its normal vector.
Together with one observation $\mathbf{P}$ with coordinates $\{\mathbf{p}_i\}_{i=1}^N$, these per-node feature vectors $\mathbf{f}_i$ are used for feature metric learning of matrix $\mathbf{M}$ as described in Section~\ref{sec:learning}.

The inter-node sample difference square $d_{i,j}$ in (\ref{eq:optimize_c_constraint}) now denotes the squared Euclidean distance between the coordinates of $\mathbf{p}_i$ and $\mathbf{p}_j$, namely,
\begin{equation}
   d_{i,j} = \|\vec{p}_i-\vec{p}_j\|_2^2,~~i \sim j,
   \label{eq:euclidean_dist}
\end{equation}
where $i \sim j$ denotes $\mathbf{p}_i$ and $\mathbf{p}_j$ are corresponding points that are connected. 

A flowchart of the proposed feature graph learning for point cloud denoising is demonstrated in Fig.~\ref{fig:flowchart}, and an algorithmic summary is presented in Algorithm~\ref{alg:denoising}.

\begin{algorithm}[htb]
    \caption{3D Point Cloud Denoising}
    \label{alg:denoising}
    \SetKwInOut{Input}{~~Input}\SetKwInOut{Output}{Output}
    \Input{Noisy point cloud $\mathbf{P} \in \mathbb{R}^{N \times 3}$, number of patches $M$, number of nearest neighbors $k$, number of nearest patches $\varepsilon$, trace constraint $C$, optimization parameter $\gamma$}
    \Output{Denoised point cloud $\mathbf{Y}$}
    Initialize $\mathbf{Y}$ with $\mathbf{P}$; \\
    \For{$iter=1,2,...$}{
        $[n_{x}^{i}, n_{y}^{i}, n_{z}^{i}]_{i=1}^{N} \gets$ estimate normal for $\mathbf{Y}$; \\
        Downsample $\mathbf{Y}$ via uniform sampling, denoted as $\mathbf{C}=\{ c_i \}_{i=1}^{M}$; \\
        Initialize $M$ empty patches $\mathbf{V}$; \\
        \For{$i=1,...,M$}{
            $\mathbf{V}_i \gets$ the nearest $k$ points from $c_i$ in $\mathbf{Y}$;
        }
        \For{$i=1,...,M$}{
            $\mathcal{B}_i \gets$ the nearest $\varepsilon$ patches to $\mathbf{V}_i$; \\
            Find corresponding node pairs from $\mathcal{B}_i$ via the algorithm in Sec.~\ref{subsec:graph_construction}; \\
        }
        Initialize $\mathbf{M}$ with a diagonal matrix with diagonal entries $C/6$; \\
        $d_{i,j} \gets$ Euclidean distance via \eqref{eq:euclidean_dist} for each node pair $(i,j)$; \\
        $g_{i,j} \gets$ feature distance for each node pair $(i,j)$; \\
        \Repeat{\eqref{eq:optimize_c_constraint} converges}{
            Solve off-diagonal entries of $\mathbf{M}$ via Block Coordinate Descent algorithm in Sec.~\ref{subsec:off_diag_entries}; \\
            Solve diagonal entries of $\mathbf{M}$ via Proximal Descent algorithm in Sec.~\ref{subsection:opt_diag}; \\
        }
        Compute adjacency matrix $\mathbf{W}$ over all patches; \\
        Compute Laplacian matrix $\mathbf{L}$; \\
        Construct selection matrix $\mathcal{T}$; \\
        $\mathbf{Y} \gets$ Solve \eqref{eq:sol};
    }
    
\end{algorithm}


\section{Experimental Results}
\label{sec:results}
\begin{figure*}[htbp]
    \centering
    \subfigure[Noisy input (161K points)]{\includegraphics[width=0.19\textwidth]{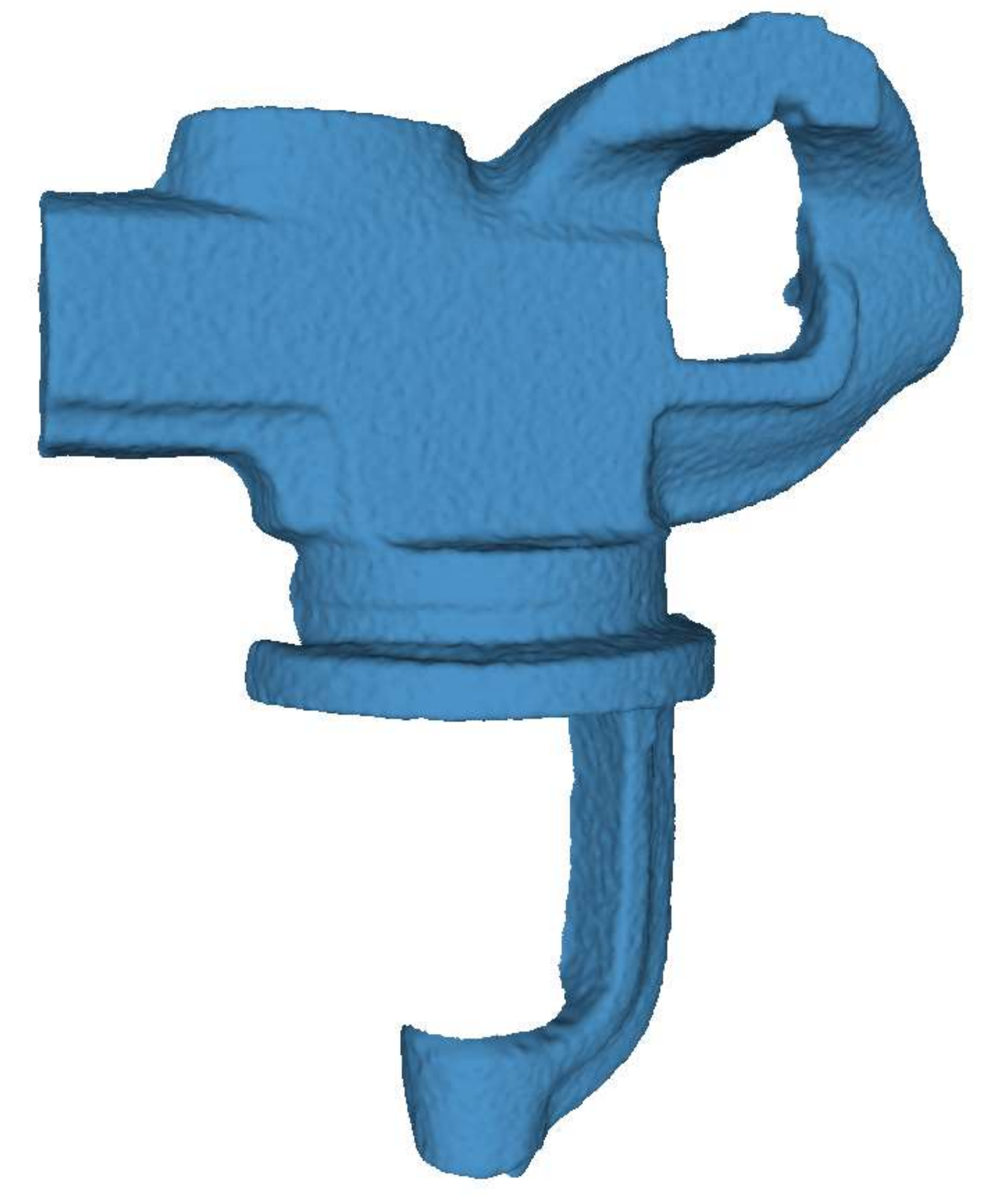}}
 	\subfigure[Iteration 1]{\includegraphics[width=0.19\textwidth]{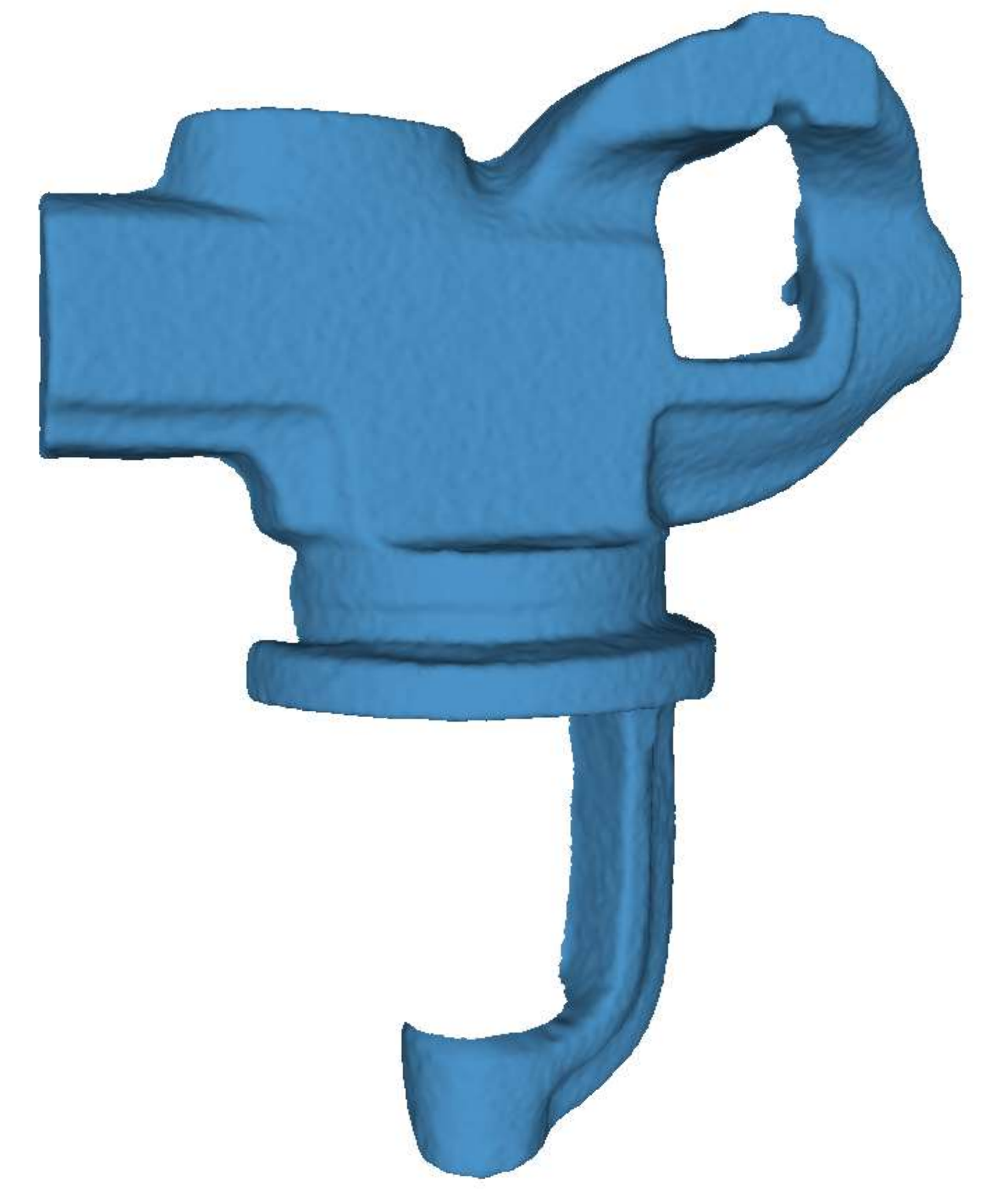}}
 	\subfigure[Iteration 3]{\includegraphics[width=0.19\textwidth]{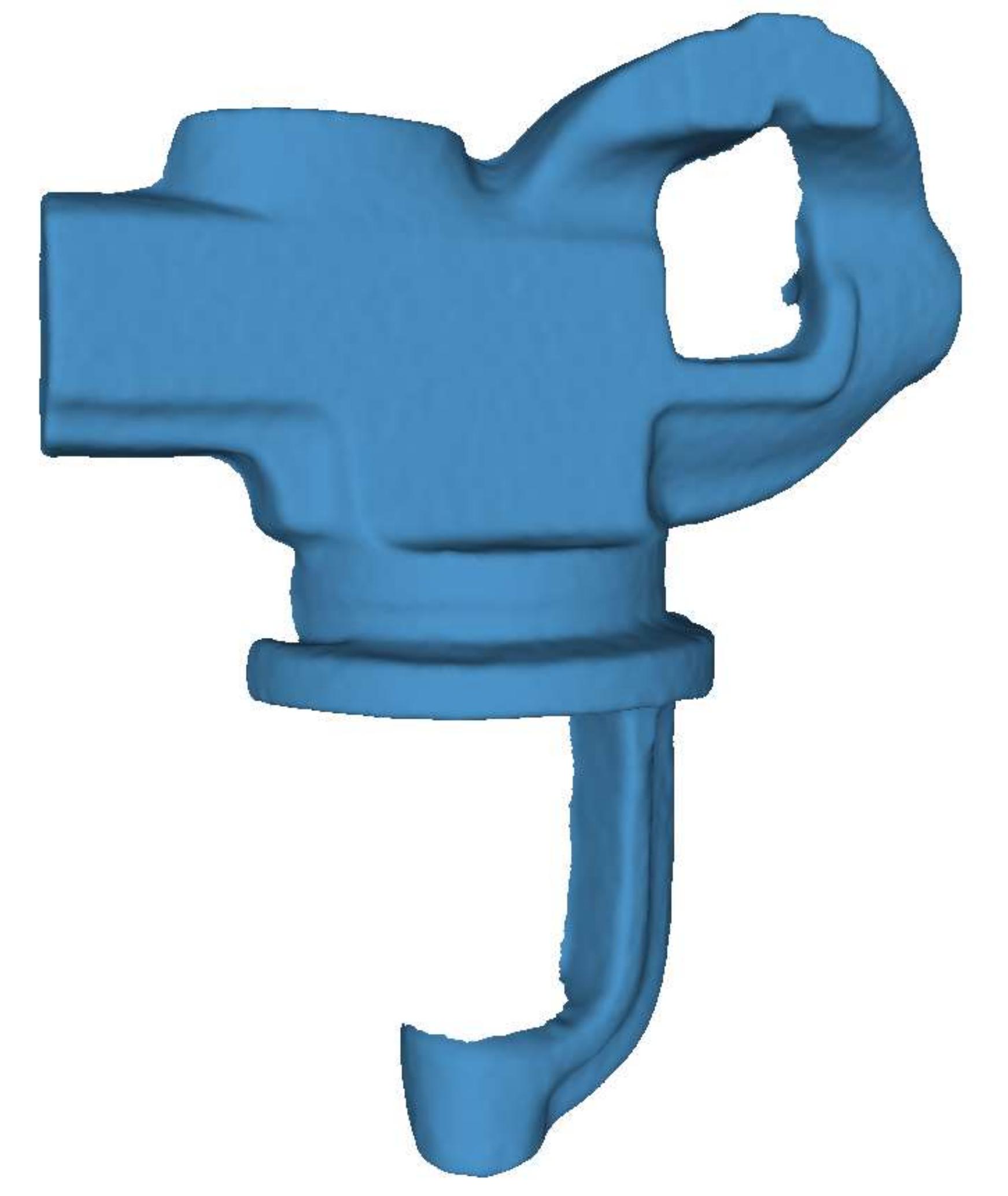}}
 	\subfigure[Iteration 5]{\includegraphics[width=0.19\textwidth]{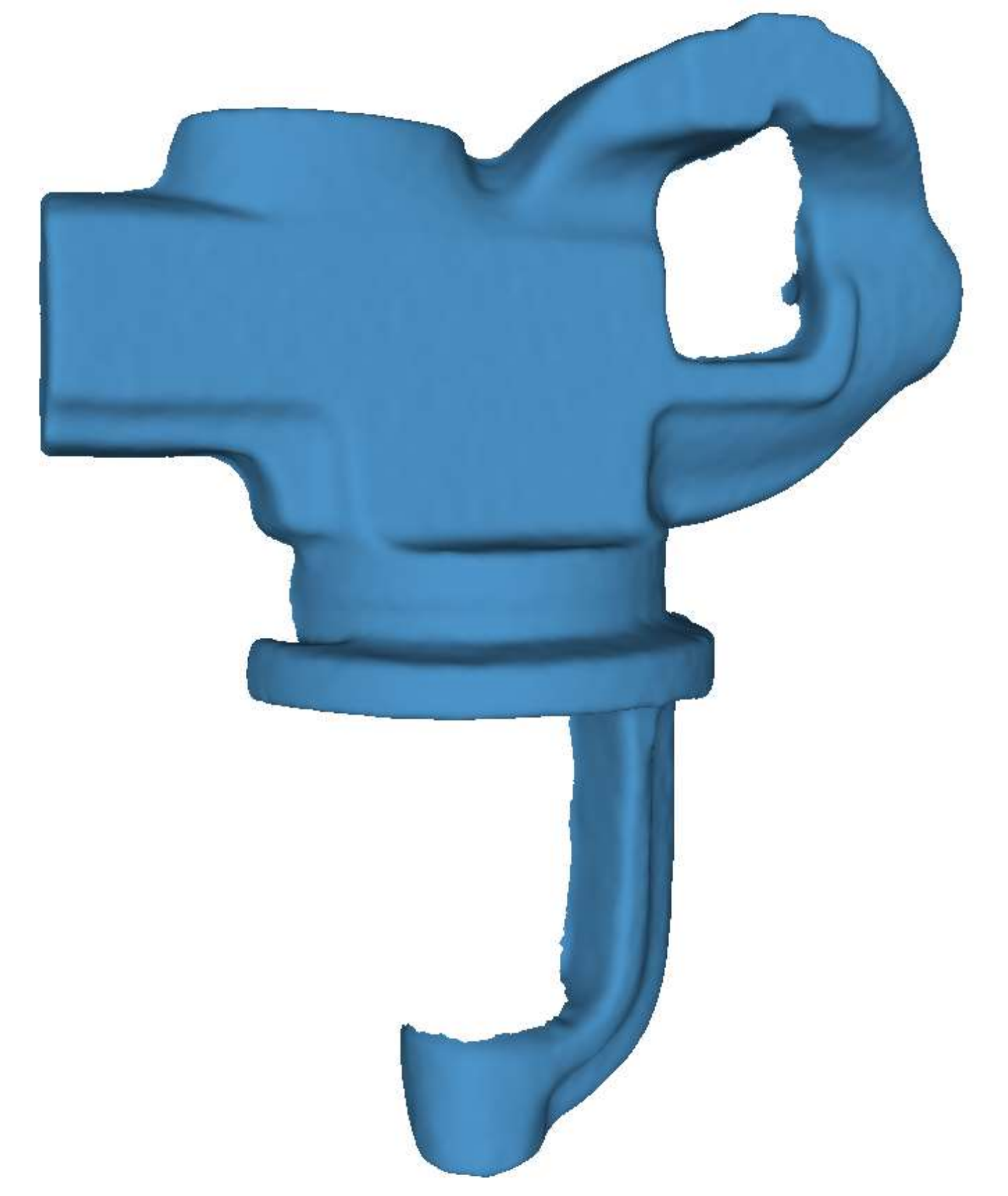}}
 	\subfigure[Iteration 7]{\includegraphics[width=0.19\textwidth]{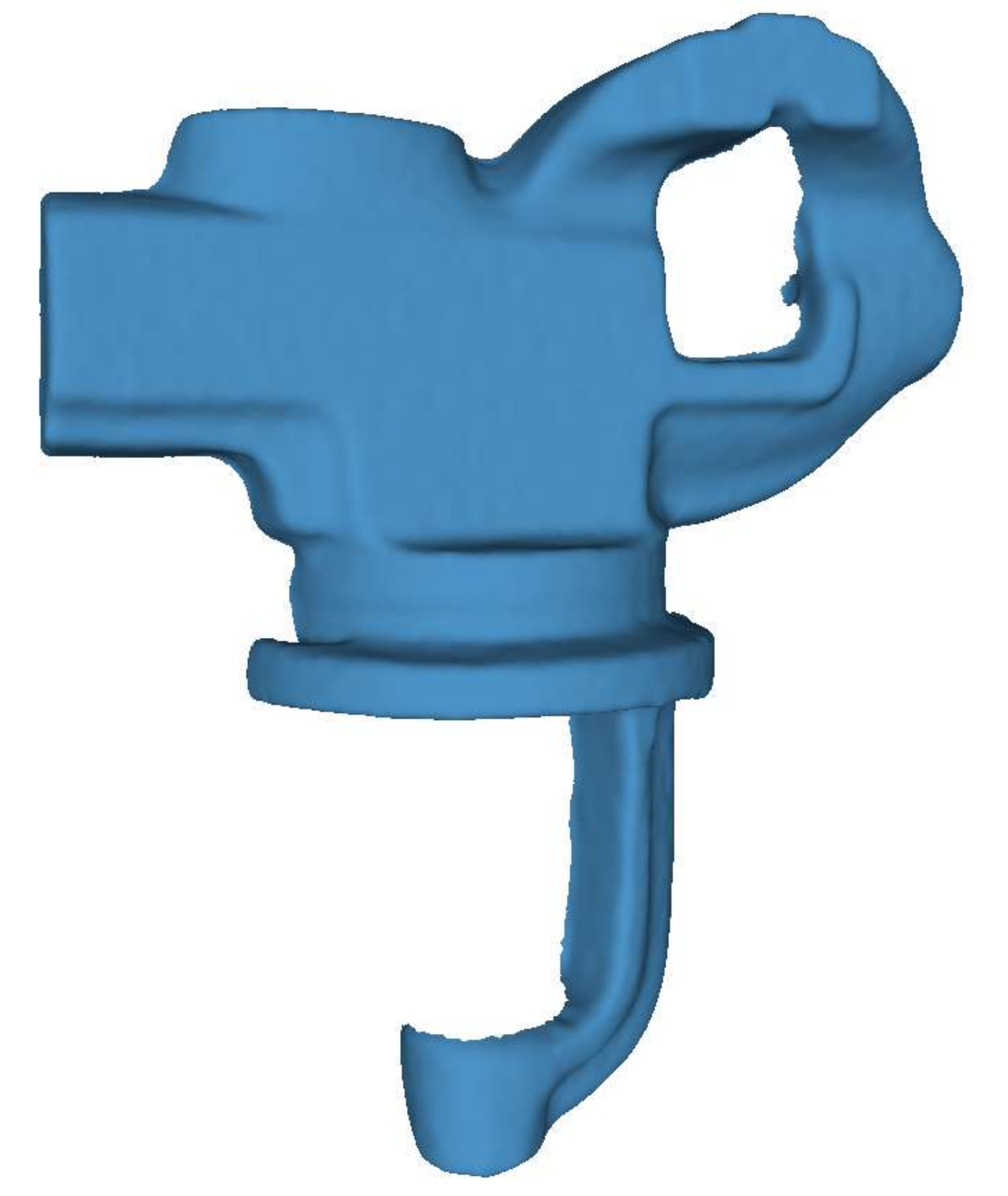}}
    \caption{Results on \texttt{Iron Vise} model in different iterations with real-world noise obtained by a laser scanner.}
    \label{fig:iter_results}
\end{figure*}

\begin{figure*}[htbp]
    \centering
    \subfigure[Ground-truth]{\includegraphics[width=0.135\textwidth]{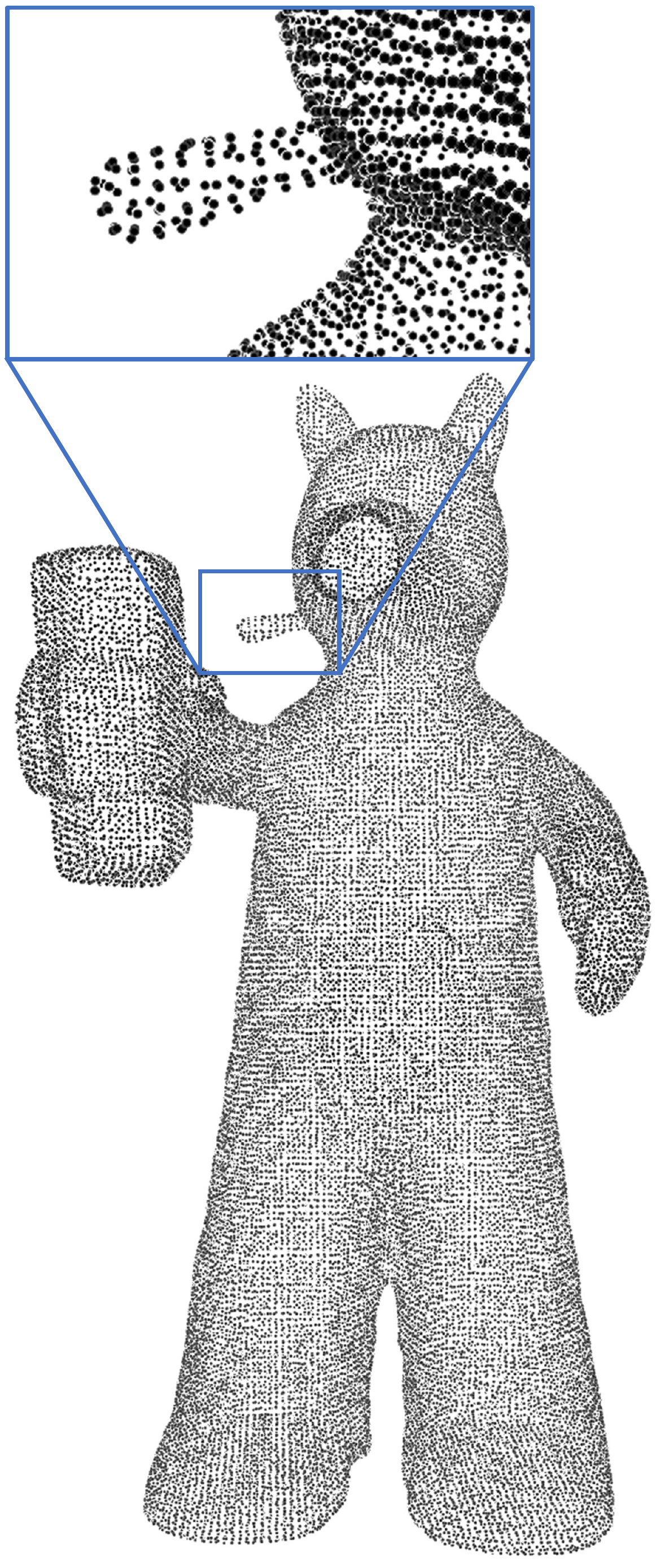}}
    \subfigure[Noisy]{\includegraphics[width=0.135\textwidth]{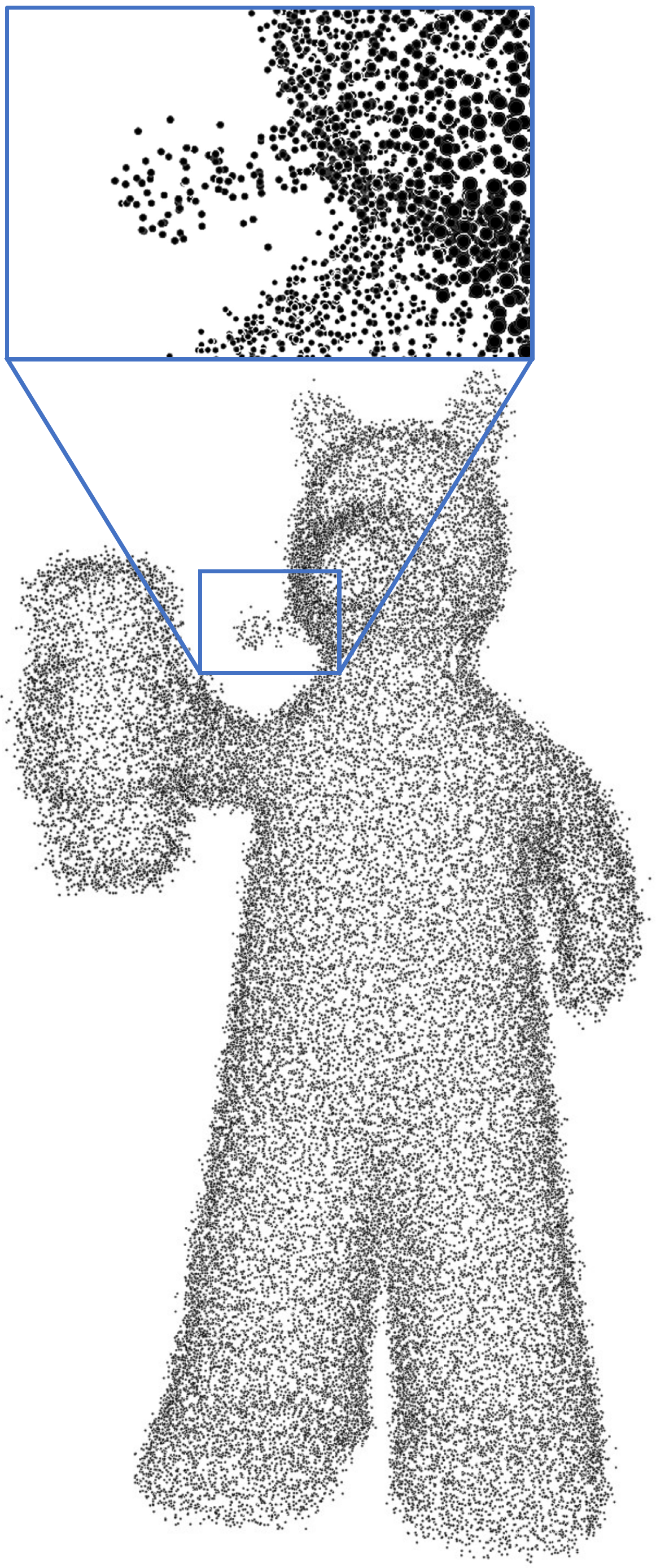}}
    \subfigure[APSS]{\includegraphics[width=0.135\textwidth]{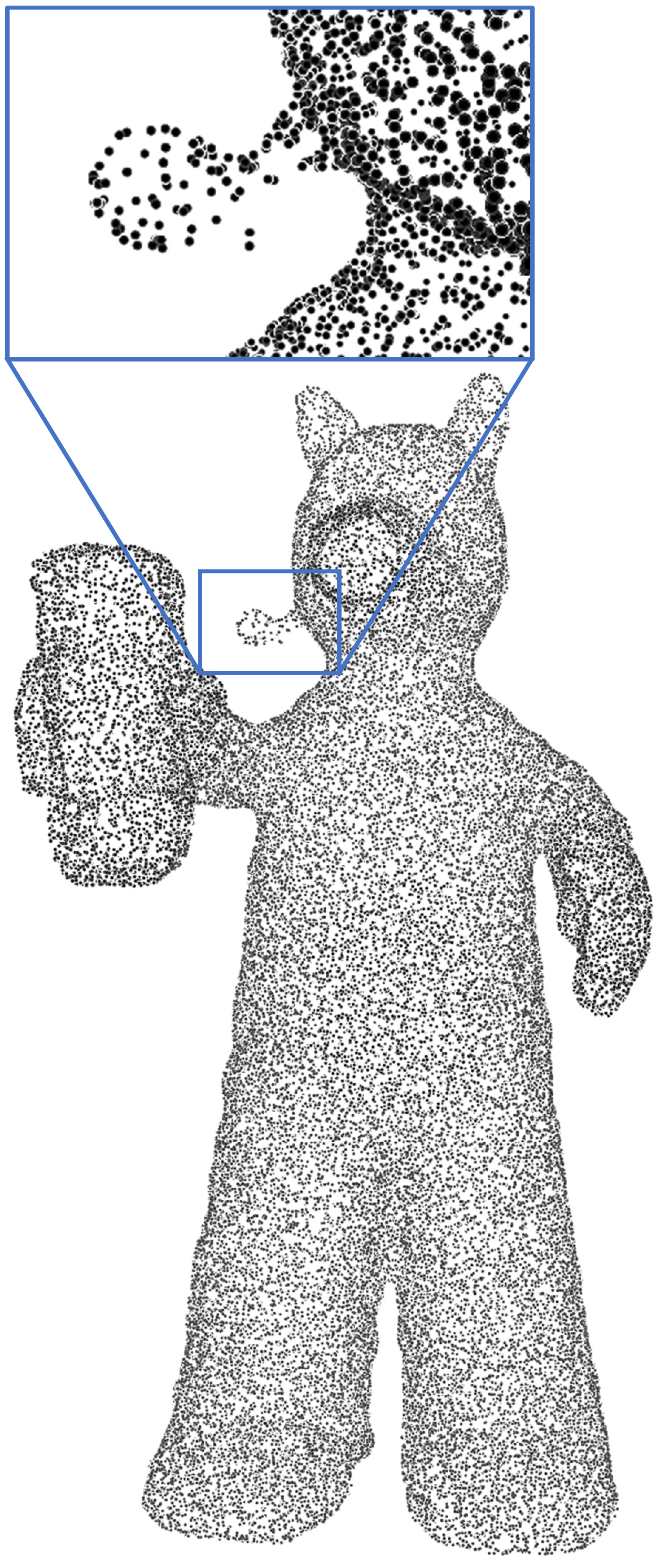}}
    \subfigure[AWLOP]{\includegraphics[width=0.135\textwidth]{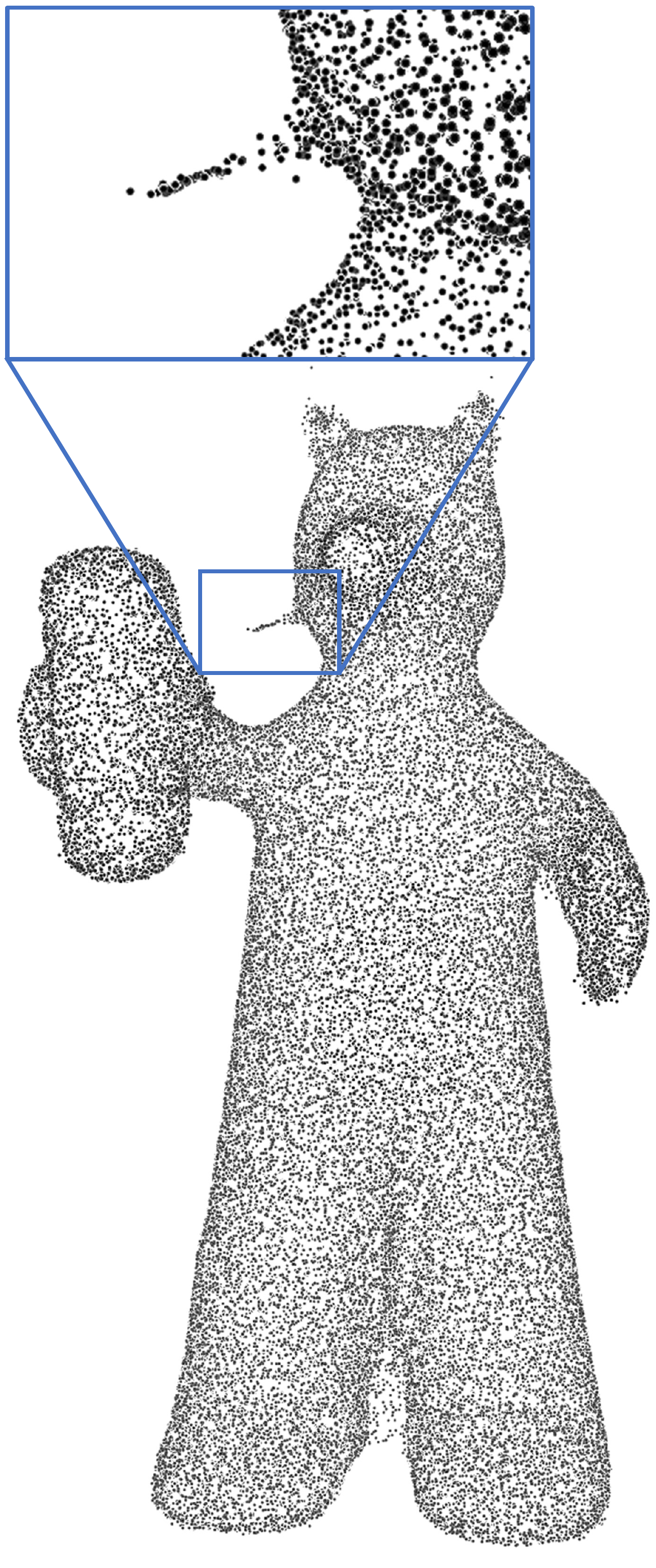}}
    \subfigure[MRPCA]{\includegraphics[width=0.135\textwidth]{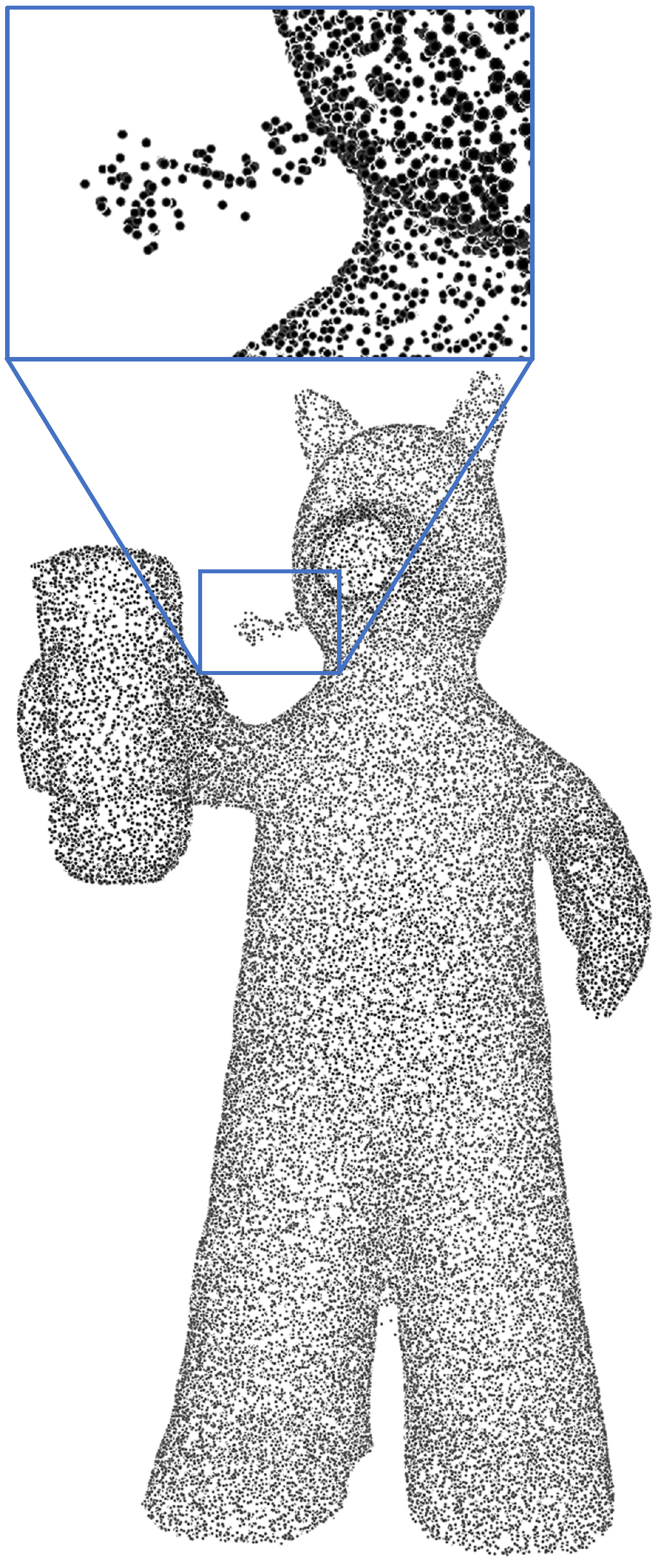}}
    \subfigure[GLR]{\includegraphics[width=0.135\textwidth]{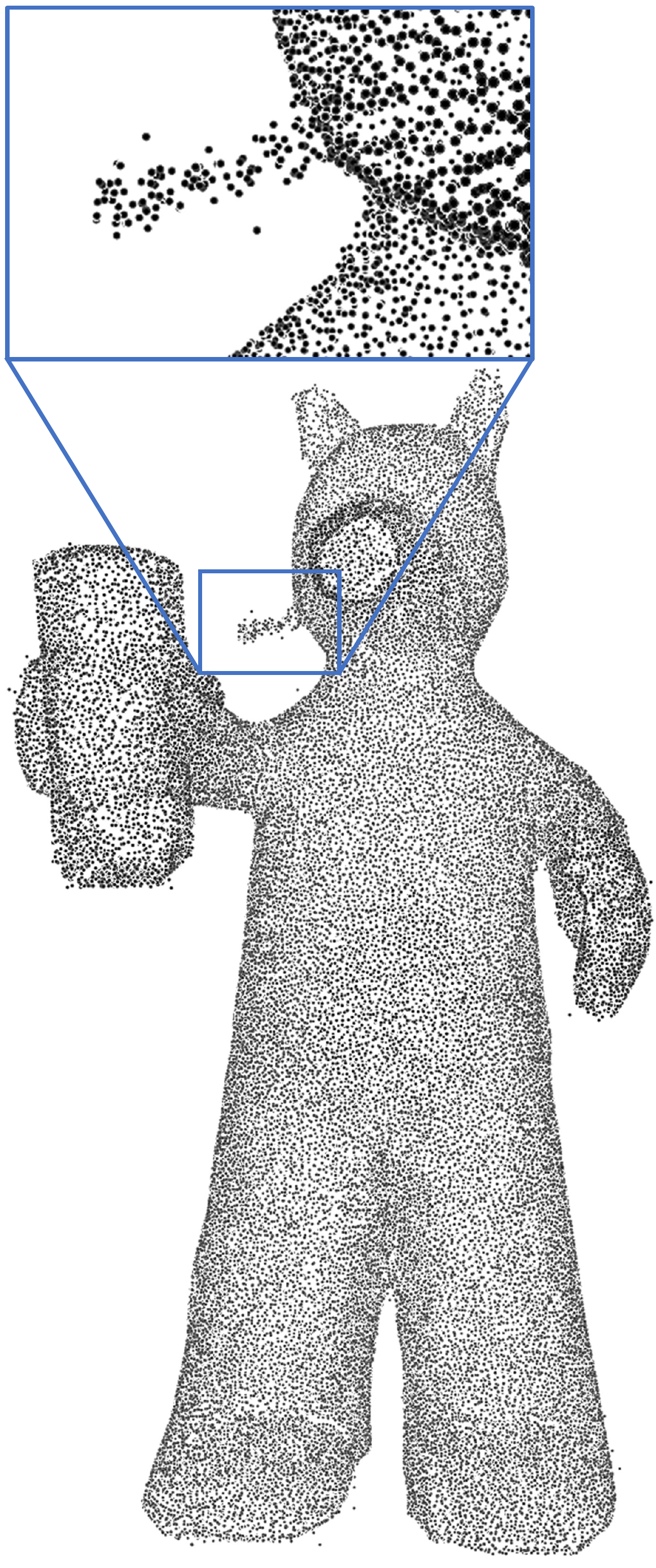}}
    \subfigure[Ours]{\includegraphics[width=0.135\textwidth]{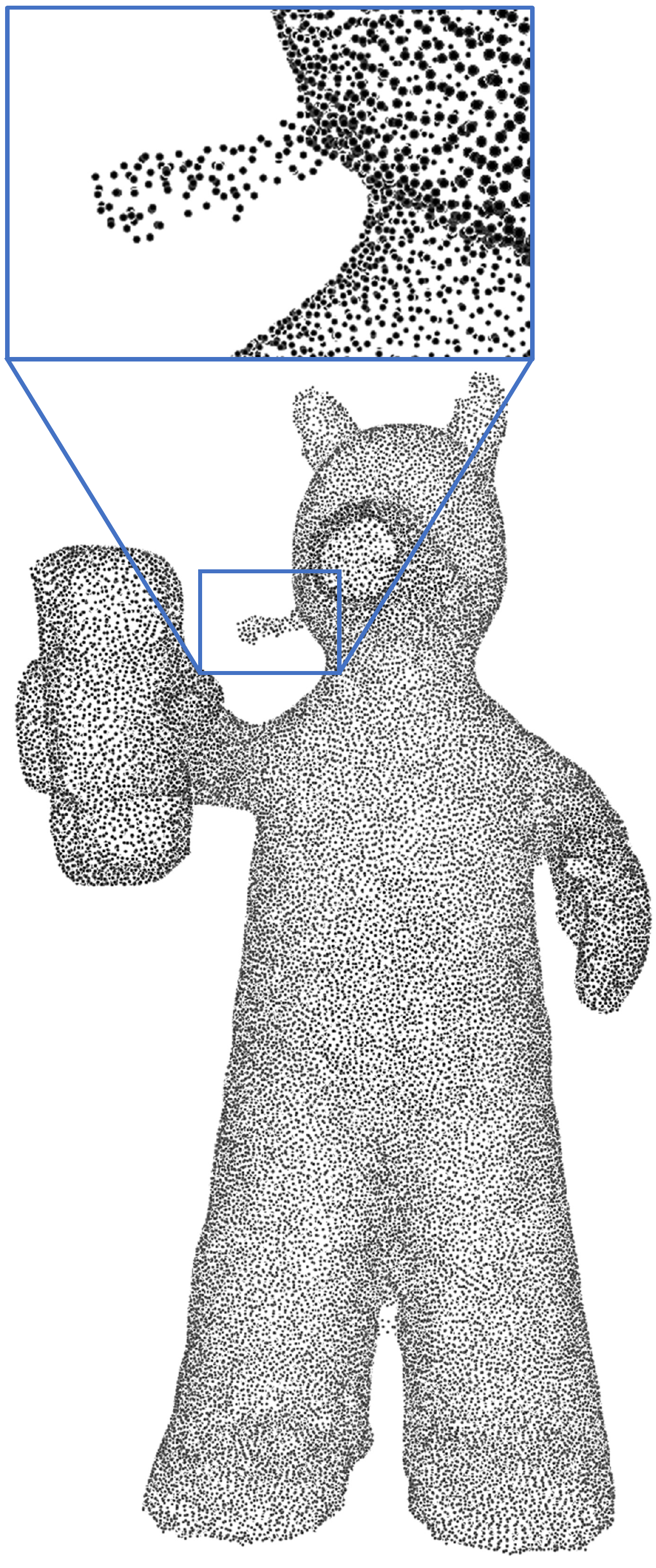}}
    \caption{Comparison results with Gaussian noise $\sigma=0.04$ for \texttt{Quasimoto}: (a) The ground truth; (b) The noisy point cloud; (c) The denoised result by APSS; (d) The denoised result by AWLOP; (e) The denoised result by MRPCA; (f) The denoised result by GLR; (g) The denoised result by our algorithm.}
    \label{fig:pointset1}
\end{figure*}

\begin{table}[]
\centering
\caption{Comparison results under Gaussian noise $\sigma=0.05$ and sampling rate $20\%$ in \texttt{Benchmark}.}
\label{tab:my-table}
\begin{tabular}{|c|c|c|c|c|c|}
\hline
 & \textbf{Anchor} & \textbf{Daratech} & \textbf{DC} & \textbf{Gargoyle} & \textbf{Quasimoto} \\ \hline \hline
 \multicolumn{6}{|c|}{MSE} \\ \hline
\textbf{Ours} & 0.240 & 0.301 & 0.223 & 0.256 & 0.197 \\ \hline
\textbf{Diagonal} & 0.251 & 0.339 & 0.241 & 0.275 & 0.205 \\ \hline \hline
 \multicolumn{6}{|c|}{SNR} \\ \hline
\textbf{Ours} & 48.17 & 43.78 & 47.04 & 46.93 & 47.87 \\ \hline
\textbf{Diagonal} & 47.72 & 42.59 & 46.23 & 46.20 & 47.50 \\ \hline
\end{tabular}
\end{table}


\subsection{Experimental Setup}

We evaluate feature graph learning for point cloud denoising by testing on several point cloud datasets: 
1) \texttt{Face} model \cite{wlop09} and \texttt{Iron Vise} model \cite{awlop13}, which are raw scans from laser scanners exhibiting \textit{real-world} noise but without ground truth; 
2) the surface reconstruction \texttt{benchmark} models including five point clouds \cite{berger2013benchmark} and \texttt{Fandisk} model \cite{rosma13}, which are clean point clouds as ground truth. 
We add Additive White Gaussian Noise (AWGN) to \texttt{benchmark} with a range of standard deviation $\sigma=\{0.02, 0.03, 0.04, 0.05, 0.10\}$ for extensive objective comparison. 
AWGN with $\sigma=0.005$ is added to \texttt{Fandisk} for subjective comparison.  
For the choice of $C$ in \eqref{eq:optimize_c_constraint}, we first select a median value of $3$, and search around this value at a step size of 0.1 to obtain good results for each dataset. 
We observe that the denoising performance of our algorithm is relatively insensitive to $C$ experimentally.

We compare the proposed approach with seven competing point cloud denoising algorithms, including two MLS-based methods APSS \cite{guennebaud07} and RIMLS \cite{ouml10}, one LOP-based method AWLOP \cite{awlop13}, one sparsity based method MRPCA \cite{mattei}, two non-local based methods NLD \cite{deschaud10} and LR \cite{sarkar2018structured}, and one graph-based method GLR \cite{zeng18arxiv}. 
The implementation details are as follows. We employ the toolbox of the MeshLab software \cite{cignoni2008meshlab} to run APSS and RIMLS. 
We try each filter scale in the range $[3,10]$ and choose one with the best result.
For AWLOP, we use its function in the EAR software \cite{awlop13}, and choose the repulsion force in $\{0.3,0.4,0.5\}$ and the filter iteration in the range $[2,5]$ to acquire one best solution.
The source codes of MRPCA and GLR are provided by the authors. We try data fitting iterations in the range $[2,6]$ for MRPCA, and follow the parameter settings in \cite{zeng18arxiv} for GLR.
We implement NLD and LR in MATLAB, and follow the default settings in their papers.
The \textit{Diagonal} approach is implemented by ourselves in MATLAB.

Further, to validate the necessity of optimizing edge weights, we compare against two \textit{Baseline} schemes of our method: 1) \textit{Baseline1}, where the edge weights are \textit{randomly} set in range $[0,1]$ instead of optimizing via feature graph learning; 2) \textit{Baseline2}, where the edge weights are calculated from feature vectors using an exponential kernel.

\subsection{Experimental Results}


\subsubsection{Demonstration of Iterations}
To show the fast convergence speed of our algorithm, we demonstrate the denoising results of \texttt{Iron Vise} model in every two iterations. 
We set the weighting parameter of GLR $\gamma=0.2^i \cdot (e-1)^{1-i}$ in \eqref{eq:formulation_final}, where $e$ is the natural logarithmic base and $i$ is the iteration index starting from $1$.
$\gamma$ decreases with iterations so as to prevent over-smoothing. 
As presented in Fig.~\ref{fig:iter_results}, as the number of iterations increases, the point cloud gradually becomes smoother, until it almost converges at iteration 7.

\subsubsection{Objective Comparison}
We measure the quality of denoised results for \texttt{Benchmark} models by the Mean Squared Error ($\texttt{MSE}$) and Signal-to-Noise Ratio ($\texttt{SNR}$) between each denoised point cloud and the ground truth as in \cite{duan2018weighted}. Numerical results are listed in Table~\ref{table:mse} and Table~\ref{table:snr}, respectively.

\begin{table*}[htbp]
\caption{\texttt{MSE} comparison for different models in \texttt{Benchmark} with Gaussian noise.}
\centering
\begin{tabularx}{0.98\textwidth}{|c|Y|Y|Y|Y|Y|Y|Y|Y|Y|Y|Y|Y|}
\hline
\scriptsize{\textbf{Model}} & \scriptsize{\textbf{Noisy}} & \scriptsize{\textbf{APSS}} & \scriptsize{\textbf{RIMLS}} & \scriptsize{\textbf{AWLOP}} & \scriptsize{\textbf{NLD}} & \scriptsize{\textbf{MRPCA}} & \scriptsize{\textbf{LR}} & \scriptsize{\textbf{GLR}} & \scriptsize{\textbf{Diagonal}} & \scriptsize{\textbf{Baseline1}} & \scriptsize{\textbf{Baseline2}} & \scriptsize{\textbf{Ours}} \\ \hline \hline
\multicolumn{13}{|c|}{$\sigma=0.02$} \\ \hline
Anchor    & 0.259 & 0.208 & 0.212 & 0.237 & 0.231 & 0.202 & 0.228 & \textbf{0.189} & 0.199 & 0.197 & 0.198 & 0.194 \\ \hline
Daratech  & 0.245 & 0.203 & 0.209 & 0.228 & 0.222 & 0.225 & 0.213 & 0.197 & 0.198 & 0.196 & 0.195 & \textbf{0.192} \\ \hline
DC        & 0.237 & 0.186 & 0.198 & 0.211 & 0.206 & 0.189 & 0.206 & 0.177 & 0.180 & 0.180 & 0.180 & \textbf{0.177} \\ \hline
Gargoyle  & 0.257 & 0.208 & 0.217 & 0.230 & 0.230 & 0.215 & 0.240 & 0.202 & 0.205 & 0.204 & 0.204 & \textbf{0.200} \\ \hline
Quasimoto & 0.224 & 0.171 & 0.183 & 0.196 & 0.190 & 0.171 & 0.180 & 0.162 & 0.162 & 0.163 & 0.162 & \textbf{0.161} \\ \hline
Average   & 0.244 & 0.195 & 0.203 & 0.220 & 0.215 & 0.200 & 0.213 & 0.185 & 0.189 & 0.188 & 0.188 & \textbf{0.184} \\ \hline \hline
\multicolumn{13}{|c|}{$\sigma=0.03$} \\ \hline
Anchor    & 0.322 & 0.239 & 0.244 & 0.259 & 0.265 & 0.230 & 0.246 & \textbf{0.217} & 0.225 & 0.227 & 0.225 & 0.221 \\ \hline
Daratech  & 0.303 & 0.242 & 0.258 & 0.298 & 0.258 & 0.259 & 0.252 & 0.238 & 0.243 & 0.244 & 0.244 & \textbf{0.236} \\ \hline
DC        & 0.293 & 0.210 & 0.226 & 0.257 & 0.235 & 0.211 & 0.221 & 0.203 & 0.205 & 0.207 & 0.204 & \textbf{0.200} \\ \hline
Gargoyle  & 0.318 & 0.239 & 0.252 & 0.294 & 0.262 & 0.241 & 0.257 & 0.233 & 0.233 & 0.237 & 0.233 & \textbf{0.228} \\ \hline
Quasimoto & 0.274 & 0.188 & 0.203 & 0.226 & 0.217 & 0.187 & 0.193 & 0.176 & 0.179 & 0.181 & 0.179 & \textbf{0.175} \\ \hline
Average   & 0.302 & 0.223 & 0.236 & 0.266 & 0.247 & 0.225 & 0.233 & 0.213 & 0.217 & 0.219 & 0.217 & \textbf{0.212} \\ \hline \hline
\multicolumn{13}{|c|}{$\sigma=0.04$} \\ \hline
Anchor    & 0.372 & 0.254 & 0.263 & 0.306 & 0.297 & 0.242 & 0.259 & \textbf{0.228} & 0.236 & 0.243 & 0.236 & 0.232 \\ \hline
Daratech  & 0.348 & 0.282 & 0.308 & 0.286 & 0.295 & 0.288 & 0.283 & 0.276 & 0.280 & 0.286 & 0.284 & \textbf{0.274} \\ \hline
DC        & 0.338 & 0.227 & 0.254 & 0.270 & 0.269 & 0.223 & 0.234 & 0.228 & 0.223 & 0.225 & 0.219 & \textbf{0.215} \\ \hline
Gargoyle  & 0.368 & 0.262 & 0.277 & 0.297 & 0.294 & 0.257 & 0.269 & 0.257 & 0.253 & 0.259 & 0.253 & \textbf{0.245} \\ \hline
Quasimoto & 0.318 & 0.201 & 0.219 & 0.218 & 0.252 & 0.199 & 0.204 & 0.187 & 0.195 & 0.195 & 0.191 & \textbf{0.182} \\ \hline
Average   & 0.348 & 0.245 & 0.264 & 0.275 & 0.281 & 0.241 & 0.249 & 0.235 & 0.237 & 0.242 & 0.237 & \textbf{0.229} \\ \hline \hline
\multicolumn{13}{|c|}{$\sigma=0.05$} \\ \hline
Anchor    & 0.417 & 0.267 & 0.281 & 0.315 & 0.331 & 0.253 & 0.270 & 0.244 & 0.246 & 0.248 & 0.246 & \textbf{0.240} \\ \hline
Daratech  & 0.387 & 0.350 & 0.373 & 0.359 & 0.330 & 0.325 & 0.347 & 0.308 & 0.319 & 0.326 & 0.322 & \textbf{0.301} \\ \hline
DC        & 0.381 & 0.251 & 0.265 & 0.324 & 0.306 & 0.239 & 0.247 & 0.241 & 0.231 & 0.247 & 0.245 & \textbf{0.222} \\ \hline
Gargoyle  & 0.412 & 0.292 & 0.305 & 0.365 & 0.334 & 0.277 & 0.281 & 0.273 & 0.266 & 0.274 & 0.268 & \textbf{0.256} \\ \hline
Quasimoto & 0.362 & 0.229 & 0.242 & 0.267 & 0.291 & 0.207 & 0.218 & 0.209 & 0.195 & 0.203 & 0.196 & \textbf{0.193} \\ \hline
Average   & 0.392 & 0.278 & 0.293 & 0.326 & 0.318 & 0.260 & 0.273 & 0.255 & 0.251 & 0.260 & 0.255 & \textbf{0.242} \\ \hline \hline
\multicolumn{13}{|c|}{$\sigma=0.10$} \\ \hline
Anchor    & 0.631 & 0.389 & 0.402 & 0.536 & 0.571 & 0.382 & 0.398 & 0.407 & 0.352 & 0.360 & 0.350 & \textbf{0.333} \\ \hline
Daratech  & 0.533 & 0.504 & 0.542 & 0.466 & 0.508 & 0.445 & 0.431 & 0.446 & 0.404 & 0.404 & 0.402 & \textbf{0.398} \\ \hline
DC        & 0.575 & 0.403 & 0.464 & 0.502 & 0.529 & 0.405 & 0.402 & 0.389 & 0.378 & 0.390 & 0.385 & \textbf{0.368} \\ \hline
Gargoyle  & 0.619 & 0.444 & 0.475 & 0.535 & 0.564 & 0.423 & 0.428 & 0.438 & 0.419 & 0.428 & 0.426 & \textbf{0.416} \\ \hline
Quasimoto & 0.561 & 0.402 & 0.414 & 0.473 & 0.523 & 0.388 & 0.387 & 0.356 & 0.293 & 0.312 & 0.304 & \textbf{0.286} \\ \hline
Average   & 0.584 & 0.428 & 0.459 & 0.502 & 0.539 & 0.409 & 0.409 & 0.407 & 0.369 & 0.379 & 0.373 & \textbf{0.360} \\ \hline
\end{tabularx}
\label{table:mse}
\end{table*}

\begin{table*}[htbp]
\caption{\texttt{SNR} (dB) comparison for different models in \texttt{Benchmark} with Gaussian noise.}
\centering
\begin{tabularx}{0.98\textwidth}{|c|Y|Y|Y|Y|Y|Y|Y|Y|Y|Y|Y|Y|}
\hline
\scriptsize{\textbf{Model}} & \scriptsize{\textbf{Noisy}} & \scriptsize{\textbf{APSS}} & \scriptsize{\textbf{RIMLS}} & \scriptsize{\textbf{AWLOP}} & \scriptsize{\textbf{NLD}} & \scriptsize{\textbf{MRPCA}} & \scriptsize{\textbf{LR}} & \scriptsize{\textbf{GLR}} & \scriptsize{\textbf{Diagonal}} & \scriptsize{\textbf{Baseline1}} & \scriptsize{\textbf{Baseline2}} & \scriptsize{\textbf{Ours}} \\ \hline \hline
\multicolumn{13}{|c|}{$\sigma=0.02$} \\ \hline
Anchor    & 47.41 & 49.61 & 49.41 & 48.31 & 48.53 & 49.88 & 48.69 & \textbf{50.55} & 50.03 & 50.14 & 50.11 & 50.30 \\ \hline
Daratech  & 45.85 & 47.71 & 47.44 & 46.56 & 46.82 & 46.72 & 47.27 & 48.02 & 47.96 & 48.08 & 48.12 & \textbf{48.29} \\ \hline
DC        & 46.42 & 48.83 & 48.23 & 47.59 & 47.82 & 48.68 & 47.83 & \textbf{49.34} & 49.19 & 49.16 & 49.16 & 49.33 \\ \hline
Gargoyle  & 46.91 & 49.01 & 48.57 & 48.01 & 48.01 & 48.66 & 47.57 & 49.30 & 49.18 & 49.20 & 49.19 & \textbf{49.40} \\ \hline
Quasimoto & 46.61 & 49.27 & 48.60 & 47.92 & 48.22 & 49.27 & 48.78 & 49.81 & 49.85 & 49.77 & 49.82 & \textbf{49.90} \\ \hline
Average   & 46.67 & 48.88 & 48.44 & 47.67 & 47.87 & 48.64 & 48.02 & 49.40 & 49.24 & 49.27 & 49.28 & \textbf{49.44} \\ \hline \hline
\multicolumn{13}{|c|}{$\sigma=0.03$} \\ \hline
Anchor    & 45.25 & 48.24 & 48.00 & 46.69 & 47.16 & 48.60 & 47.91 & \textbf{49.20} & 48.82 & 48.72 & 48.83 & 48.99 \\ \hline
Daratech  & 43.70 & 46.00 & 45.46 & 45.12 & 45.34 & 45.18 & 45.59 & 46.13 & 45.92 & 45.89 & 45.91 & \textbf{46.22} \\ \hline
DC        & 44.32 & 47.64 & 46.94 & 46.04 & 46.49 & 47.62 & 47.10 & 47.94 & 47.84 & 47.76 & 47.90 & \textbf{48.11} \\ \hline
Gargoyle  & 44.75 & 47.63 & 47.12 & 46.39 & 46.68 & 47.52 & 46.88 & 47.87 & 47.89 & 47.70 & 47.86 & \textbf{48.09} \\ \hline
Quasimoto & 44.58 & 48.34 & 47.57 & 46.53 & 46.89 & 48.40 & 48.09 & 49.00 & 48.83 & 48.72 & 48.82 & \textbf{49.06} \\ \hline
Average   & 44.52 & 47.57 & 47.01 & 46.15 & 46.51 & 47.46 & 47.11 & 48.02 & 47.86 & 47.76 & 47.86 & \textbf{48.09} \\ \hline \hline
\multicolumn{13}{|c|}{$\sigma=0.04$} \\ \hline
Anchor    & 43.78 & 47.60 & 47.27 & 45.74 & 46.02 & 48.09 & 47.41 & \textbf{48.67} & 48.34 & 48.04 & 48.35 & 48.51 \\ \hline
Daratech  & 42.34 & 44.46 & 43.58 & 44.32 & 43.98 & 44.25 & 44.41 & 44.64 & 44.53 & 44.30 & 44.36 & \textbf{44.73} \\ \hline
DC        & 42.86 & 46.84 & 45.71 & 45.11 & 45.15 & 47.00 & 46.54 & 46.80 & 47.03 & 46.93 & 47.21 & \textbf{47.38} \\ \hline
Gargoyle  & 43.31 & 46.69 & 46.14 & 45.44 & 45.53 & 46.88 & 46.44 & 46.89 & 47.05 & 46.81 & 47.06 & \textbf{47.37} \\ \hline
Quasimoto & 43.09 & 47.68 & 46.80 & 46.85 & 45.40 & 47.80 & 47.52 & 48.40 & 48.00 & 47.97 & 48.20 & \textbf{48.67} \\ \hline
Average   & 43.07 & 46.65 & 45.90 & 45.49 & 45.21 & 46.80 & 46.46 & 47.08 & 46.99 & 46.81 & 47.04 & \textbf{47.33} \\ \hline \hline
\multicolumn{13}{|c|}{$\sigma=0.05$} \\ \hline
Anchor    & 42.65 & 47.09 & 46.59 & 45.44 & 44.94 & 47.64 & 46.99 & 48.00 & 47.90 & 47.80 & 47.92 & \textbf{48.17} \\ \hline
Daratech  & 41.28 & 42.29 & 41.64 & 42.02 & 42.87 & 43.03 & 42.37 & 43.56 & 43.21 & 42.99 & 43.11 & \textbf{43.80} \\ \hline
DC        & 41.68 & 45.86 & 45.30 & 43.27 & 43.85 & 46.33 & 46.01 & 46.24 & 46.68 & 45.98 & 46.06 & \textbf{47.07} \\ \hline
Gargoyle  & 42.17 & 45.61 & 45.18 & 43.37 & 44.28 & 46.12 & 45.99 & 46.28 & 46.56 & 46.25 & 46.45 & \textbf{46.93} \\ \hline
Quasimoto & 41.79 & 46.36 & 45.83 & 44.83 & 43.99 & 47.39 & 46.85 & 47.28 & 47.96 & 47.55 & 47.94 & \textbf{48.08} \\ \hline
Average   & 41.91 & 45.44 & 44.91 & 43.79 & 43.99 & 46.10 & 45.64 & 46.27 & 46.46 & 46.11 & 46.30 & \textbf{46.81} \\ \hline \hline
\multicolumn{13}{|c|}{$\sigma=0.10$} \\ \hline
Anchor    & 38.52 & 43.35 & 43.04 & 40.13 & 39.51 & 43.52 & 43.12 & 42.89 & 44.33 & 44.10 & 44.38 & \textbf{44.87} \\ \hline
Daratech  & 38.10 & 38.66 & 37.92 & 39.43 & 38.57 & 39.88 & 40.24 & 39.86 & 40.84 & 40.82 & 40.88 & \textbf{41.00} \\ \hline
DC        & 37.57 & 41.14 & 39.72 & 38.91 & 38.38 & 41.07 & 41.13 & 41.45 & 41.71 & 41.39 & 41.54 & \textbf{42.00} \\ \hline
Gargoyle  & 38.12 & 41.44 & 40.78 & 39.57 & 39.03 & 41.91 & 41.79 & 41.56 & 41.97 & 41.76 & 41.82 & \textbf{42.06} \\ \hline
Quasimoto & 37.43 & 40.78 & 40.49 & 39.13 & 38.13 & 41.11 & 41.13 & 41.96 & 43.88 & 43.26 & 43.53 & \textbf{44.14} \\ \hline
Average   & 37.95 & 41.07 & 40.39 & 39.43 & 38.72 & 41.50 & 41.48 & 41.54 & 42.55 & 42.27 & 42.43 & \textbf{42.81} \\ \hline
\end{tabularx}
\label{table:snr}
\end{table*}

Both tables show that our method outperforms all the other competing approaches at various noise levels in general, especially at high noise level $\sigma=0.04$.  
Also, we outperform the \textit{Baseline} scheme with random weights, which validates optimizing edge weights is essential.  

Further, to demonstrate the case of learning from partial observation of one signal, we randomly sample a subset of points ($20\%$) in each point cloud for the learning of the feature metric $\mathbf{M}$ at noise level $\sigma=0.05$. 
On one hand, we achieve comparable results with learning from the entire single observation as listed in Table~\ref{table:mse} and Table~\ref{table:snr}. 
On the other hand, we compare with the diagonal-only method in \cite{yang2018apsipa}, where only a diagonal feature metric is learned from $20\%$ of points. 
Results show that we still outperform \cite{yang2018apsipa} in the circumstance of learning feature metric from partial observation. 
This validates the effectiveness of our method even when extending to the case of partial observation of one signal.



\subsubsection{Comparison with Vanilla Proximal Gradient}
Moreover, we compare with the na\"{i}ve realization of proximal gradient \textbf{Vanilla PG} as discussed in Section~\ref{subsec:formulation} to solve \eqref{eq:optimize_c_constraint}. 
As presented in Table~\ref{tab:naive_pg}, while our proposed algorithm approximates the original search space in \textbf{Vanilla PG} by rewriting the PD constraint, our point cloud denoising results are very close to the performance by \textbf{Vanilla PG}. 
This validates the effectiveness of our optimization approximation.

\subsubsection{Subjective Comparison}

Fig.~\ref{fig:pointset1} shows visual results of \texttt{Quasimoto} model in \texttt{Benchmark} without surface reconstruction for details. We compare with other denoising approaches at noise level $\sigma=0.04$. 
It can be seen that our results preserve structural details well, even for tiny components such as the cigarette in Fig.~\ref{fig:pointset1}. In comparison, the cigarette is distorted in all the other reconstruction results. 
Also, points in our results are more uniformly distributed than the others, even for noise with large variance.

\begin{figure*}[htbp]
    \centering
    \subfigure[Ground-truth]{\includegraphics[width=0.135\textwidth]{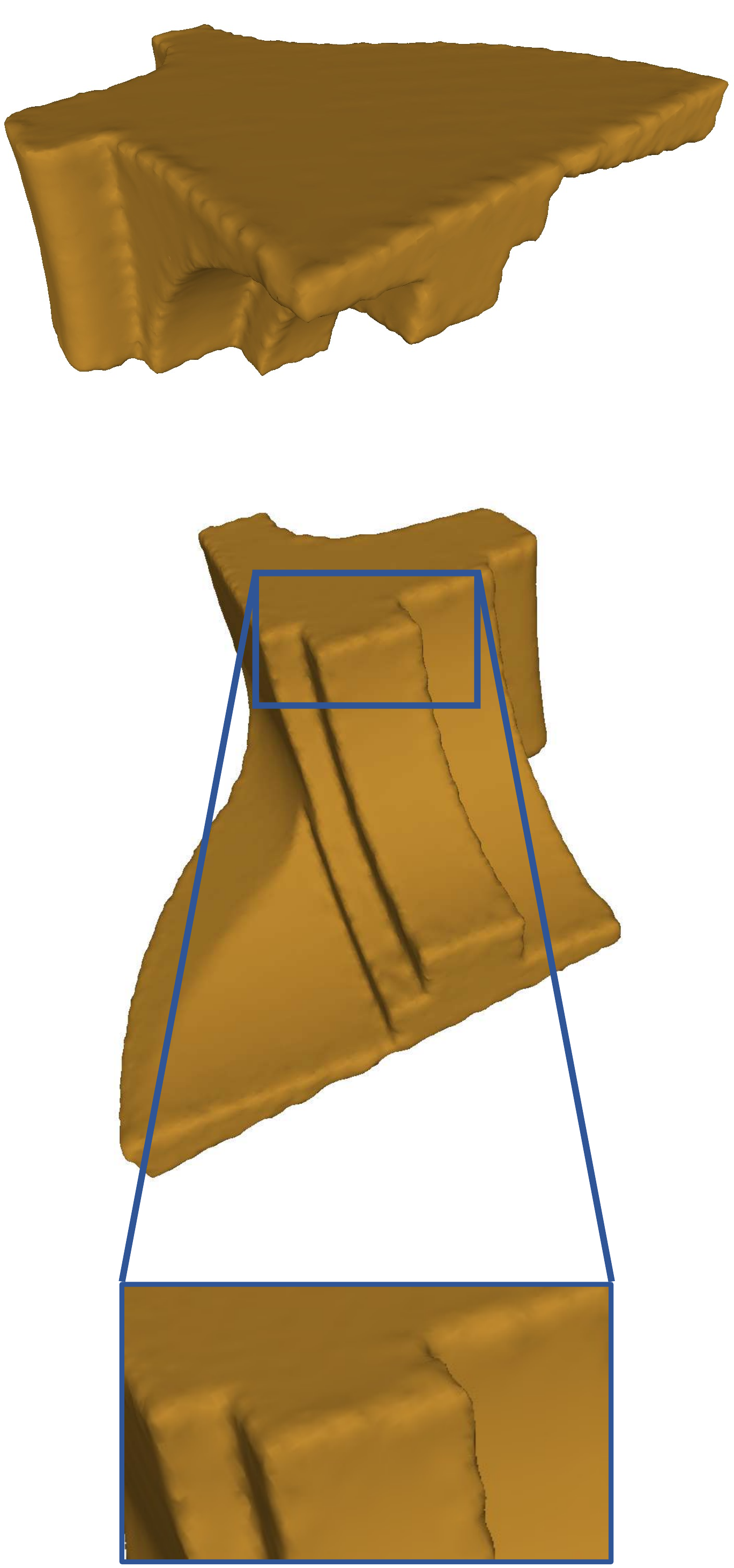}}
    \subfigure[Noisy]{\includegraphics[width=0.135\textwidth]{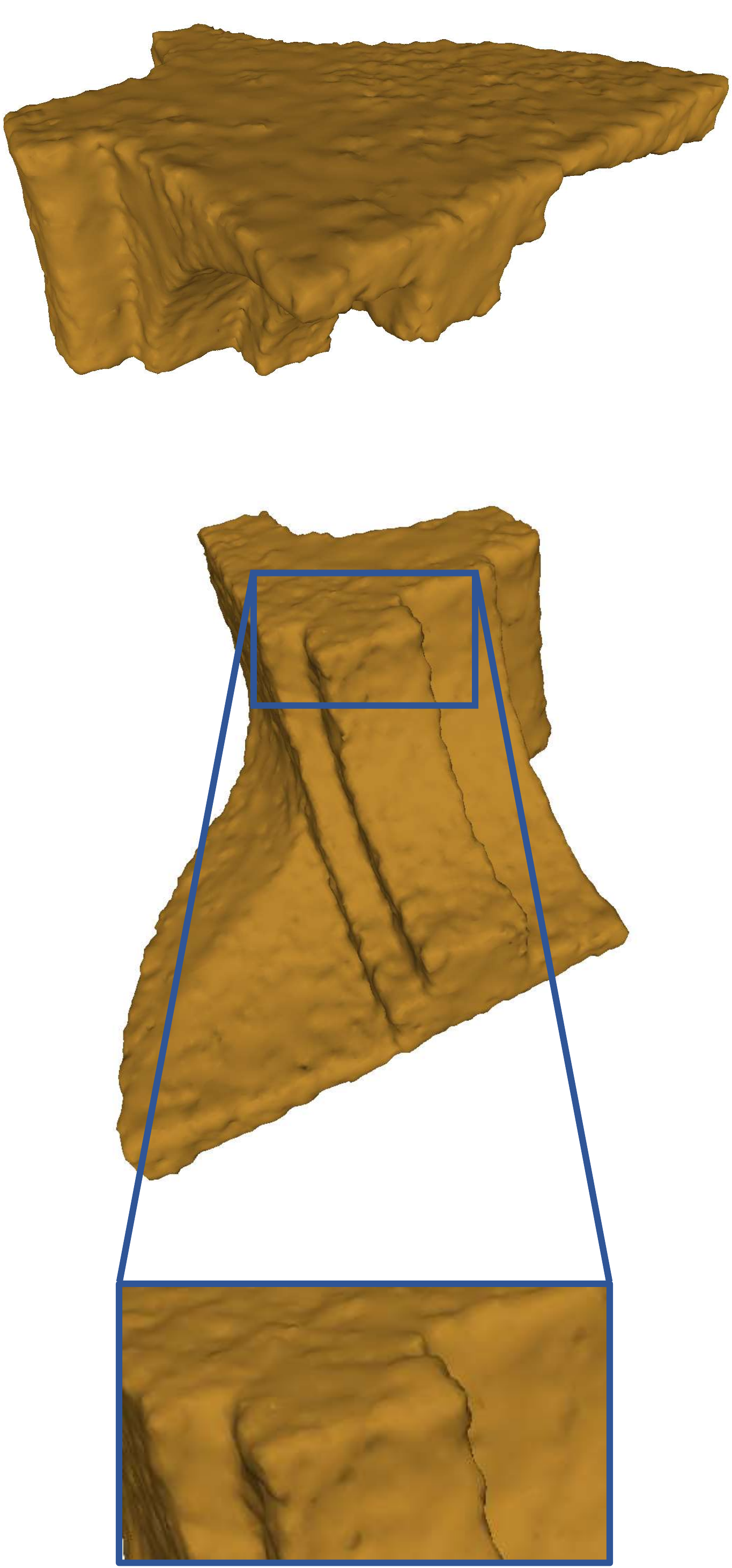}}
    \subfigure[APSS]{\includegraphics[width=0.135\textwidth]{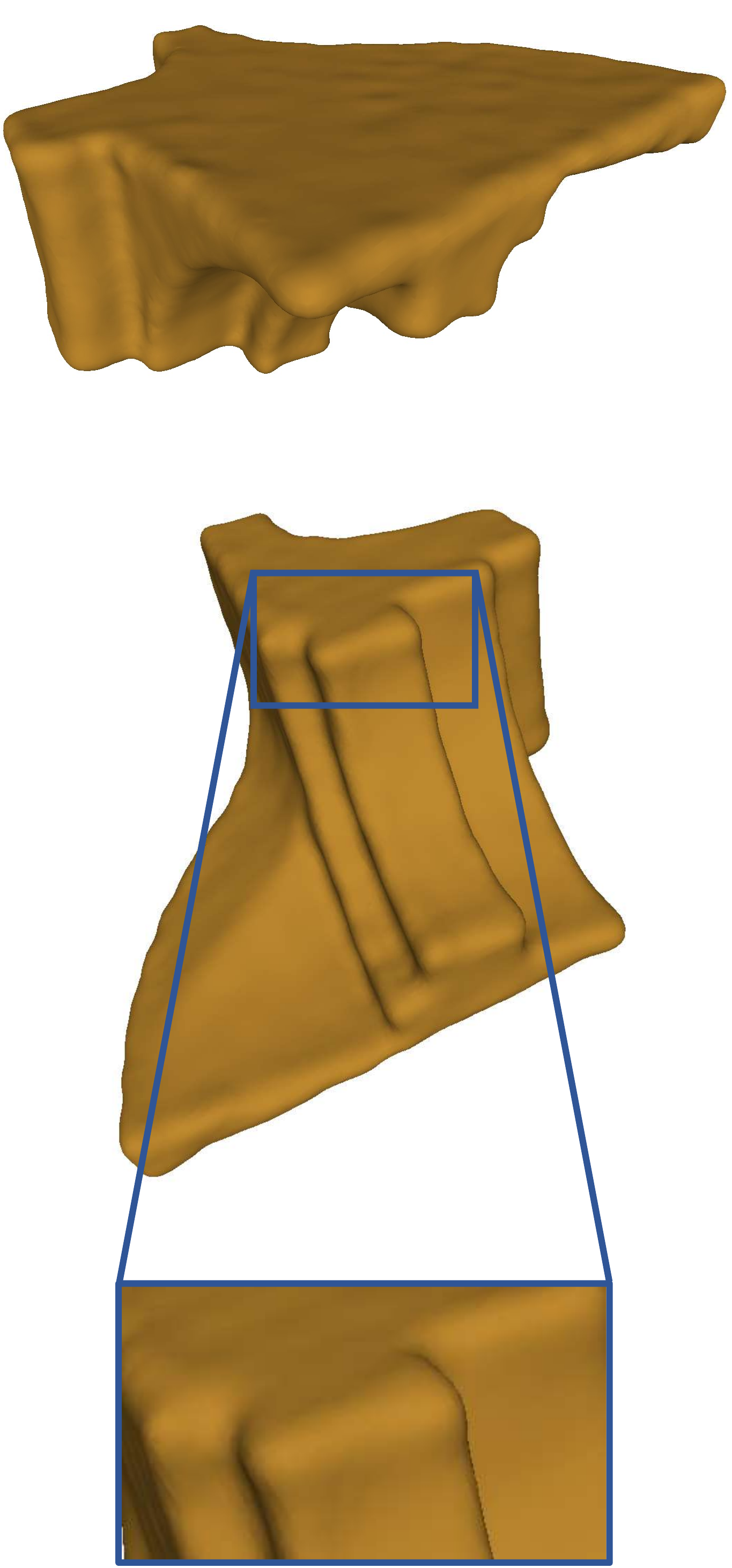}}
    \subfigure[AWLOP]{\includegraphics[width=0.135\textwidth]{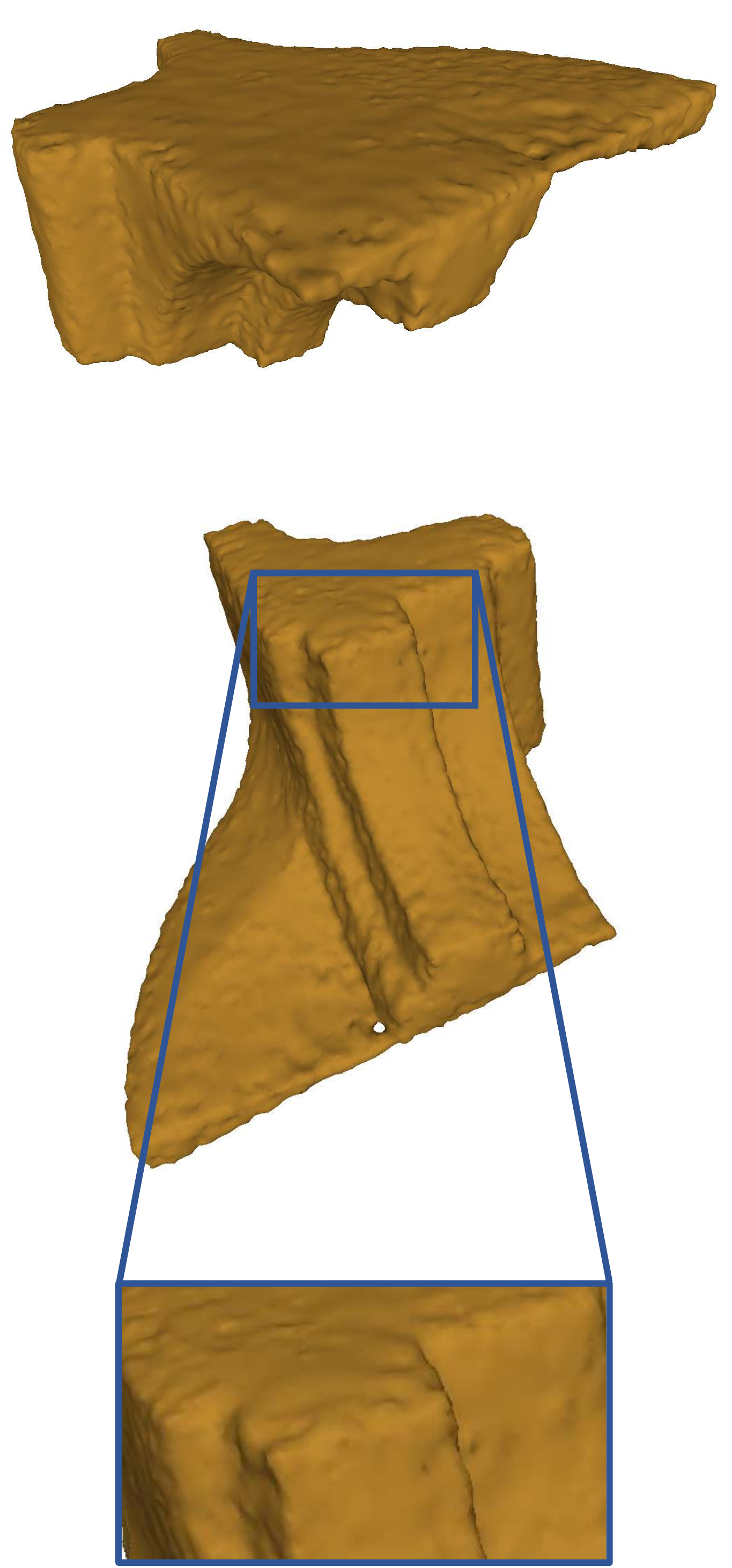}}
    \subfigure[MRPCA]{\includegraphics[width=0.135\textwidth]{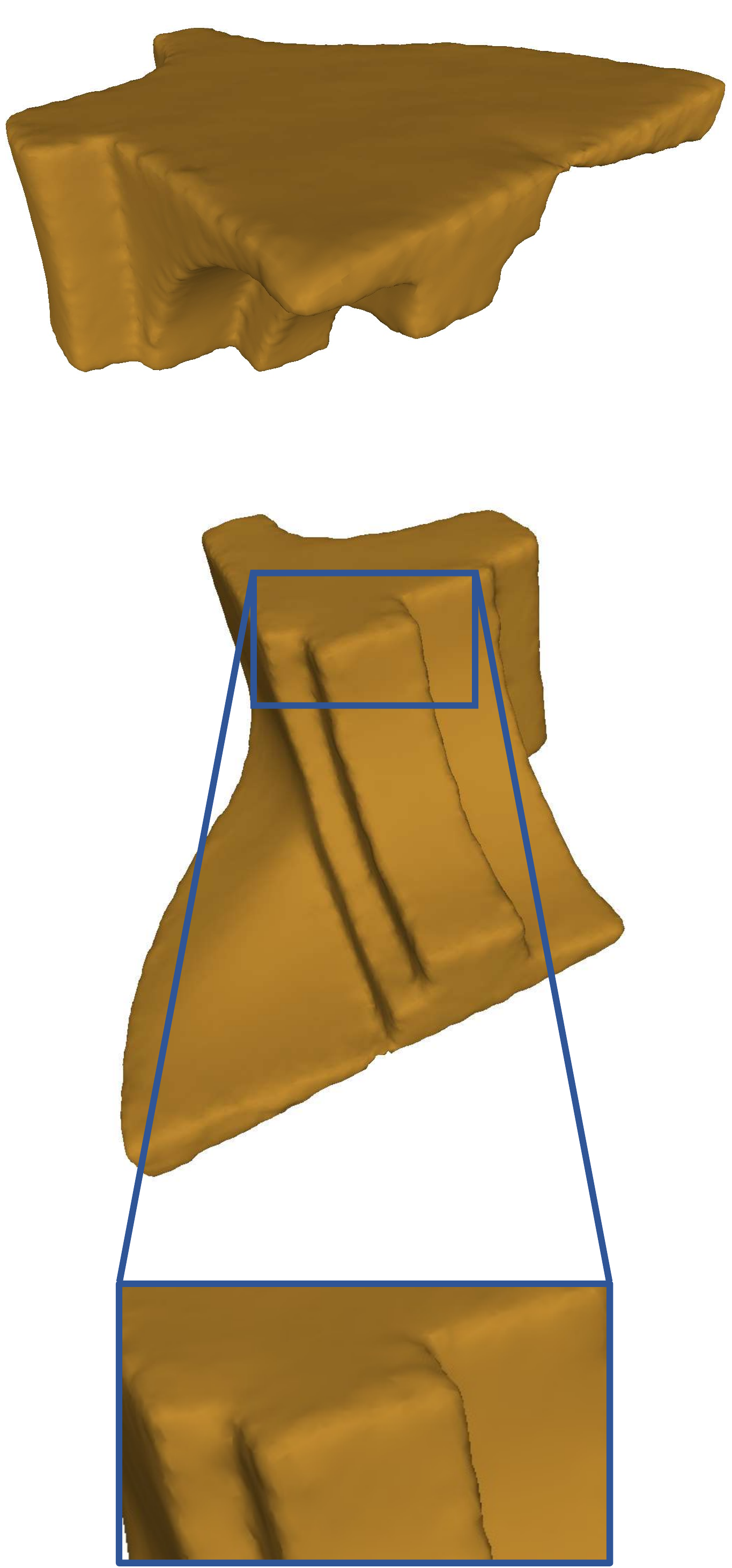}}
    \subfigure[GLR]{\includegraphics[width=0.135\textwidth]{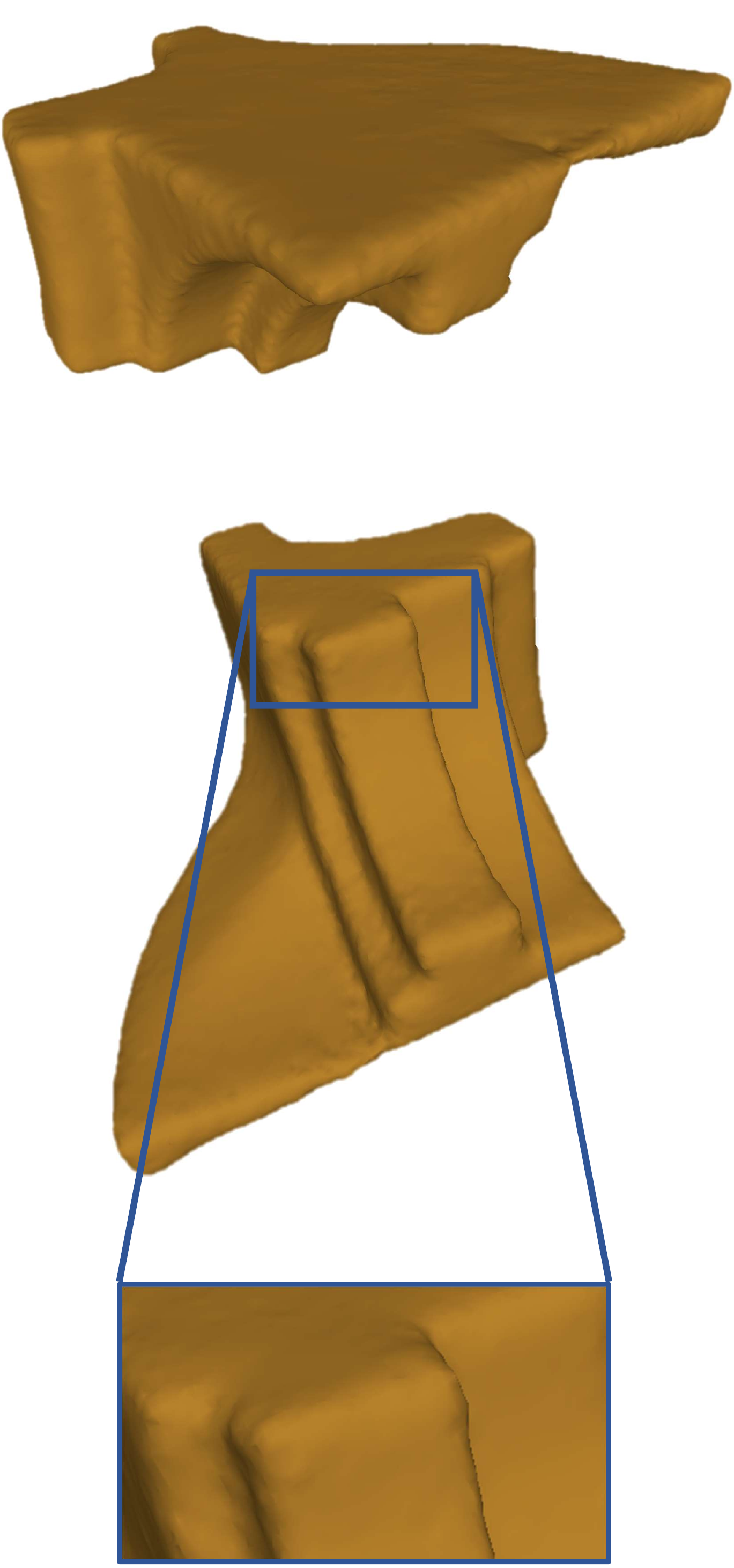}}
    \subfigure[Ours]{\includegraphics[width=0.135\textwidth]{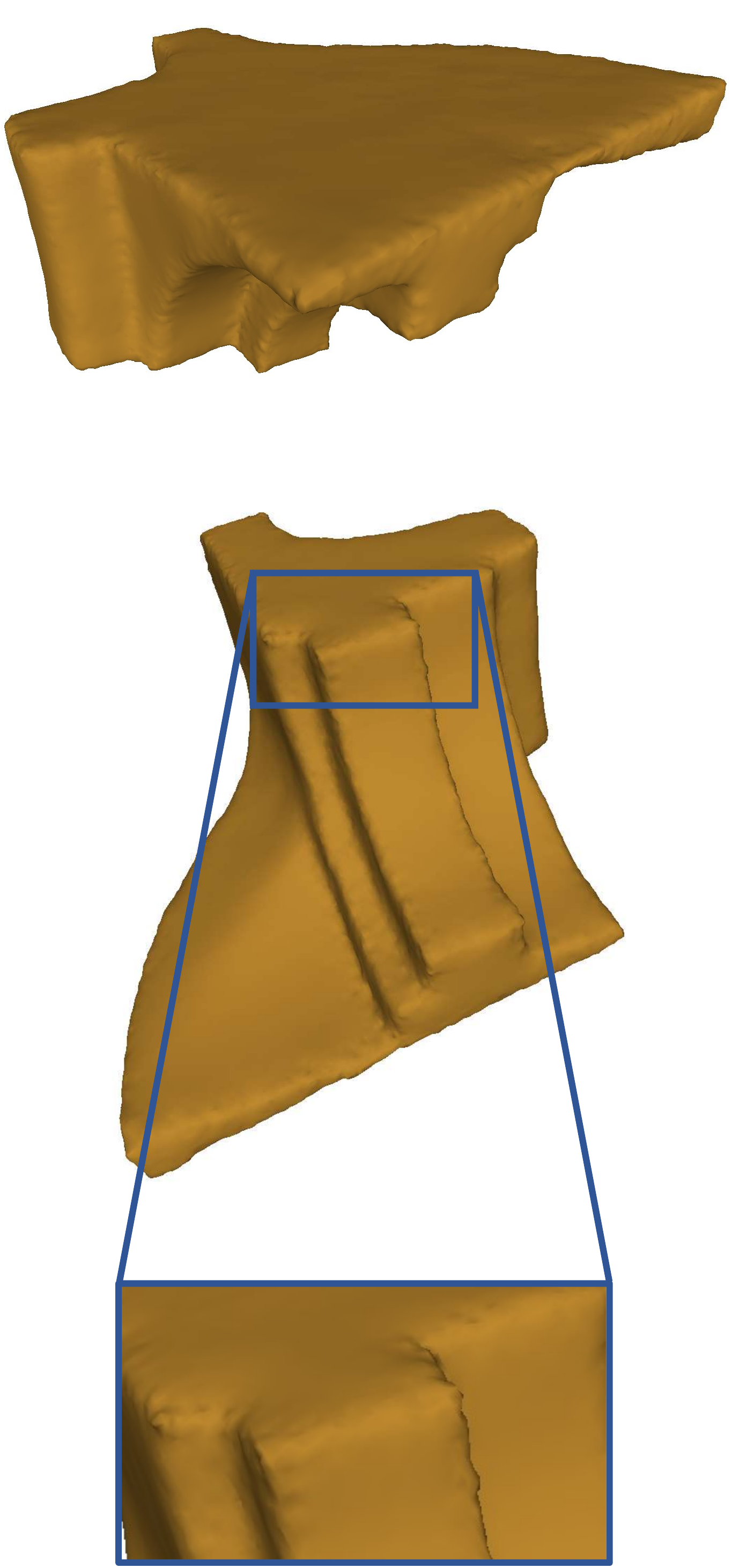}}
    \caption{Comparison results with Gaussian noise $\sigma=0.005$ for \texttt{Fandisk}, where the first row shows the bottom of \texttt{Fandisk}. (a) The ground truth; (b) The noisy point cloud; (c) The denoised result by APSS; (d) The denoised result by AWLOP; (e) The denoised result by MRPCA; (f) The denoised result by GLR; (g) The denoised result by our algorithm.}
    \label{fig:fandisk}
\end{figure*}

In Fig.~\ref{fig:fandisk}, we add AWGN to \texttt{Fandisk} model with standard deviation $\sigma=0.005$. 
We see that our method preserves sharp features, while the other approaches result in smoothed edges to various extent. Meanwhile, our method reconstructs smooth surfaces well, such as the bottom surface as presented in the first row of Fig.~\ref{fig:fandisk}.


\begin{figure}[htbp]
    \centering
    \subfigure[Noisy input (84.4K points)]{\includegraphics[width=0.45\columnwidth]{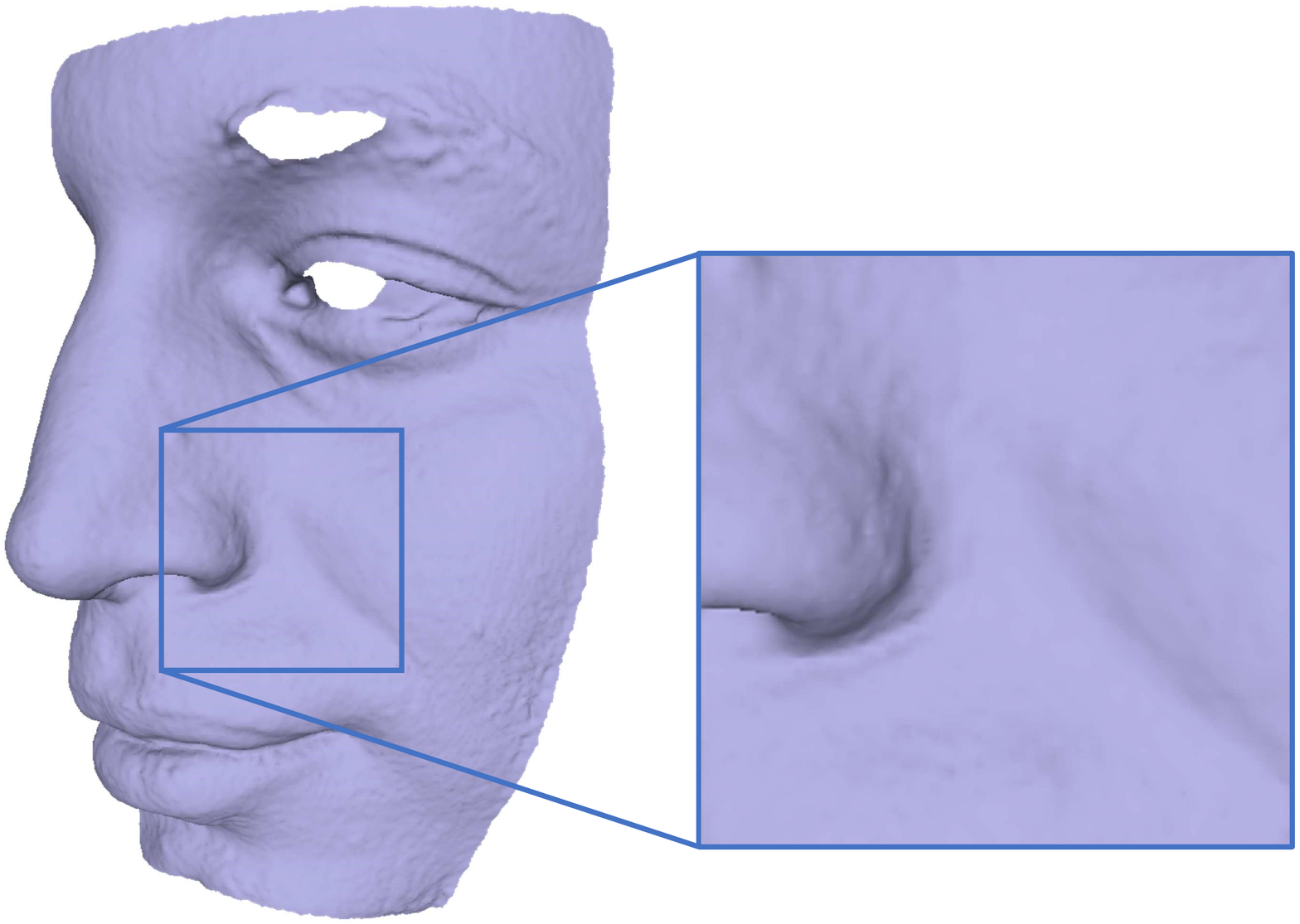}}
    \subfigure[MRPCA]{\includegraphics[width=0.45\columnwidth]{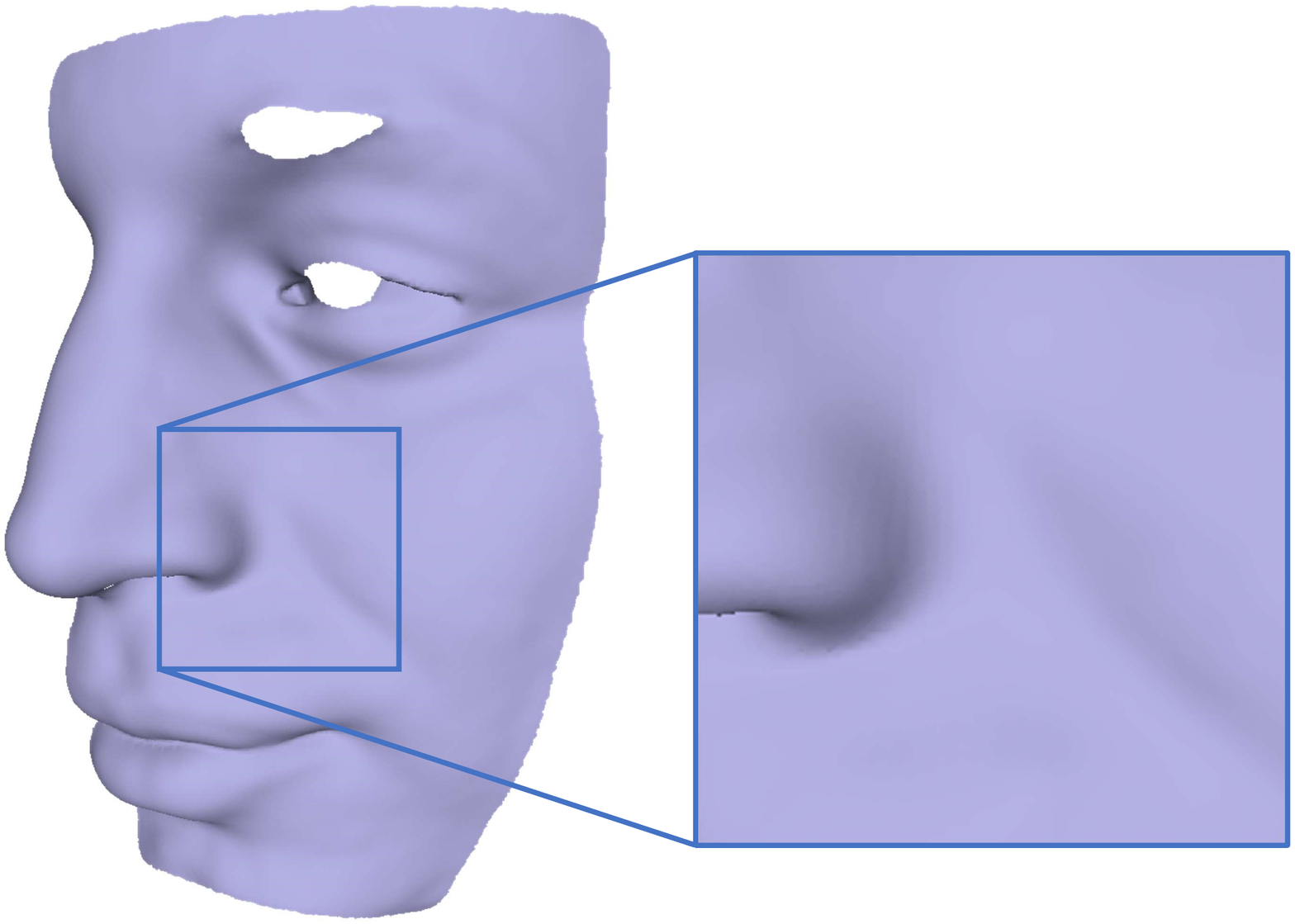}}
    \subfigure[GLR]{\includegraphics[width=0.45\columnwidth]{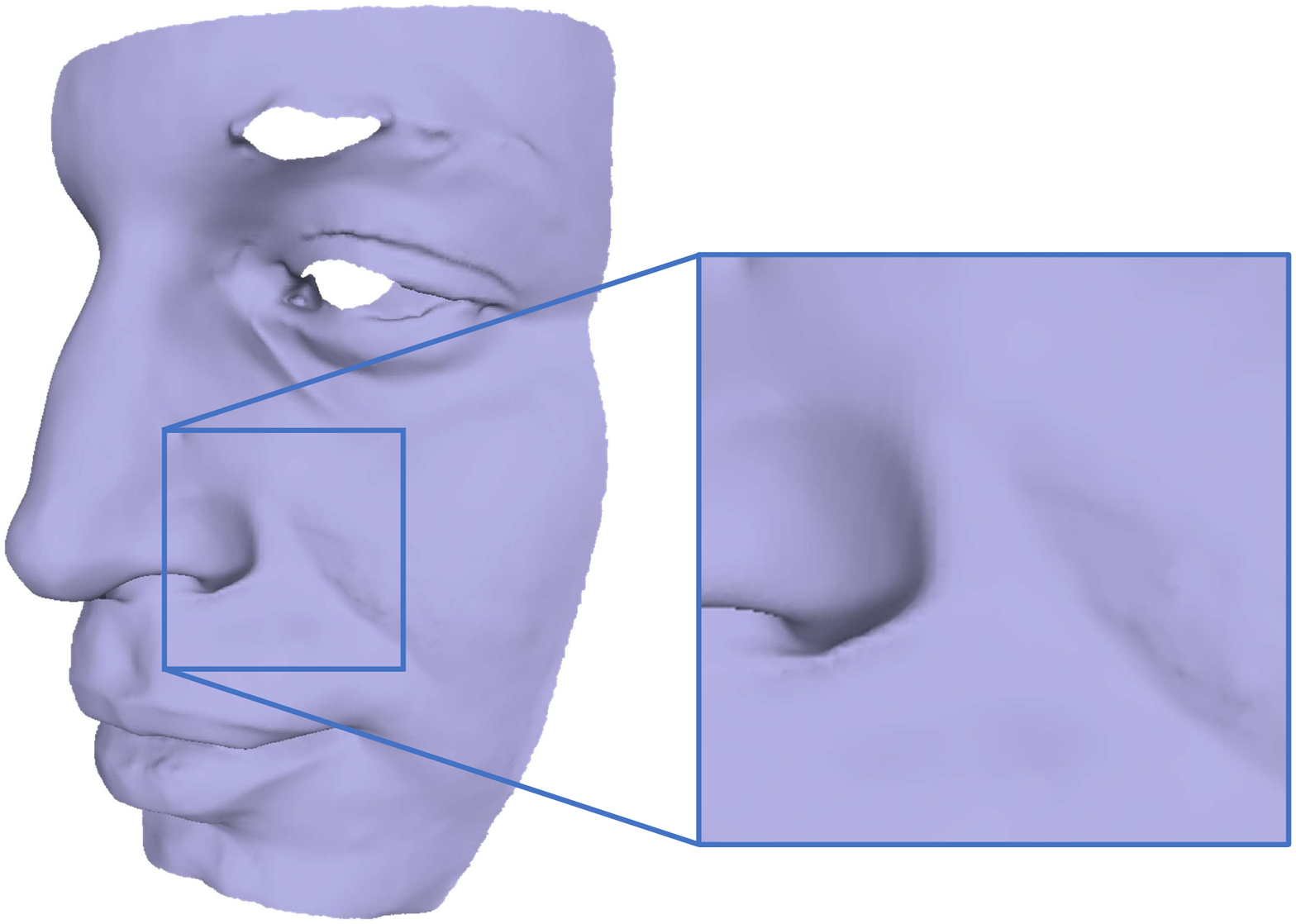}}
    \subfigure[Ours]{\includegraphics[width=0.45\columnwidth]{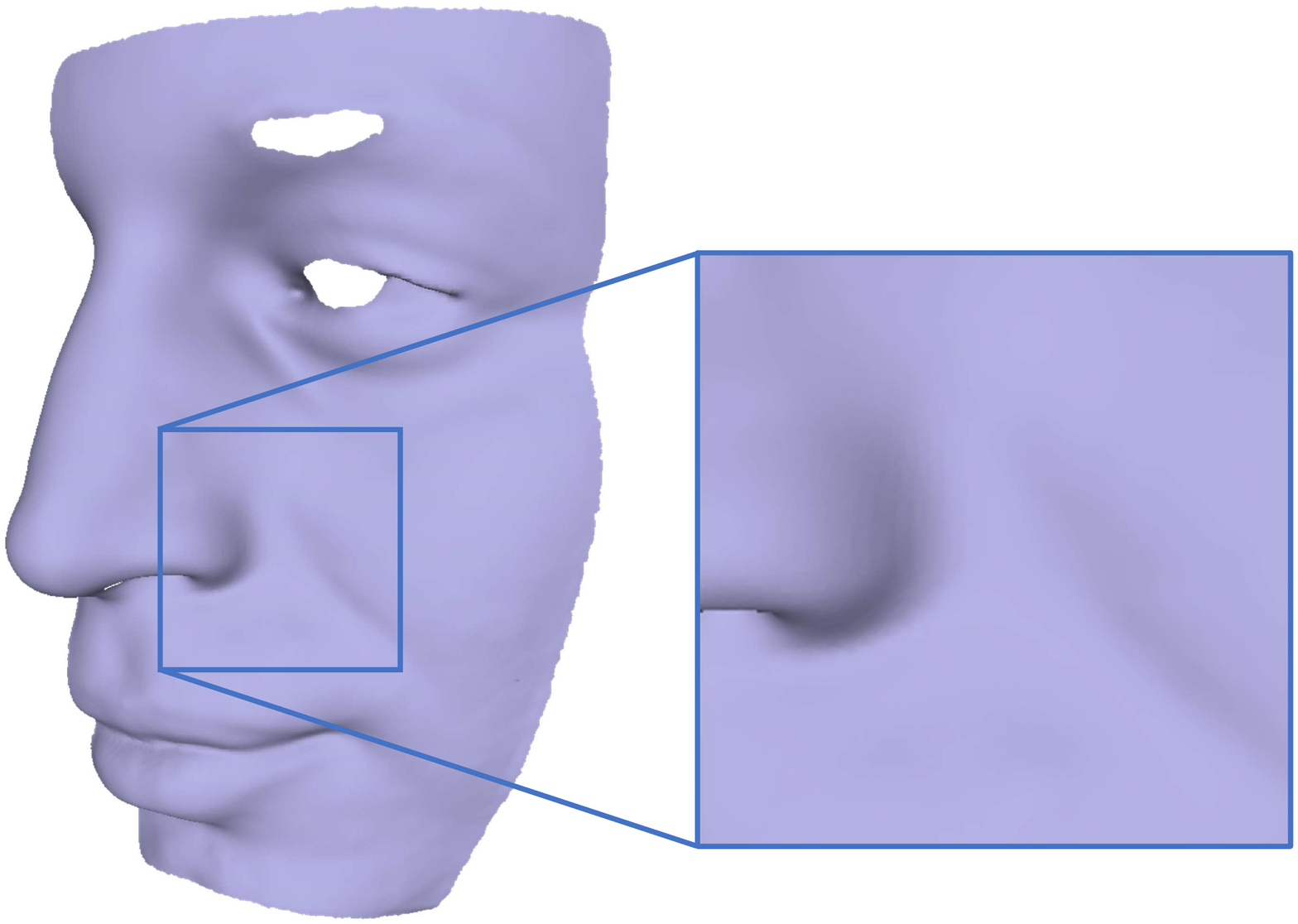}}
    \caption{Comparison results of \texttt{Face} model: (a) The real-world noisy point cloud acquired by laser scanners; (b) The denoised result by MRPCA; (c) The denoised result by GLR; (d) The denoised result by our algorithm.}
    \label{fig:face}
\end{figure}

Finally, we evaluate on \textit{real-world} noisy point clouds \texttt{Iron Vise} and \texttt{Face} acquired from laser scanners.
Typical imperfections associated with digital scans, such as noise, non-uniform point distribution, or missing data, are ubiquitous in these datasets.
As shown in Fig.~\ref{fig:iter_results} and Fig.~\ref{fig:face}, our method is able to keep local details and sharp edges while attenuating noise significantly. 

\subsection{Limitation}
The limitation of the proposed algorithm includes two aspects. 
\begin{itemize}
    \item The optimal choice of the parameter $C$ in \eqref{eq:optimize_c_constraint}---the upper bound of the trace of the feature metric $\mathbf{M}$---is not obvious. 
    We choose it empirically in the experiments, and found that our algorithm's denoising performance is relatively insensitive to $C$ experimentally. 
    \item The requirement of having a complete feature vector per graph node may not be practical for some applications.
\end{itemize}

\begin{table}[]
\centering
\caption{MSE Comparison for different optimization algorithms under Gaussian noise $\sigma=0.05$ in \texttt{Benchmark}.}
\label{tab:naive_pg}
\begin{tabular}{|c|c|c|c|c|c|}
\hline
 & \textbf{Anchor} & \textbf{Daratech} & \textbf{DC} & \textbf{Gargoyle} & \textbf{Quasimoto} \\ \hline
\textbf{Vanilla PG} & 0.240 & 0.301 & 0.222 & 0.256 & 0.191 \\ \hline
\textbf{Ours} & 0.240 & 0.301 & 0.222 & 0.256 & 0.193 \\ \hline
\end{tabular}
\end{table}

\section{Conclusion}
\label{sec:conclude}
We study feature graph learning to identify an appropriate underlying graph given a \textit{single} signal observation. 
Assuming the availability of relevant features per node, we formulate the problem as minimization of Graph Laplacian Regularizer using the Mahalanobis distance matrix $\mathbf{M}$ as variable. 
We develop a fast algorithm to alternately optimize diagonal and off-diagonal entries of $\mathbf{M}$, while keeping $\mathbf{M}$ positive definite.
We mitigate full matrix eigen-decomposition and large matrix inverse for fast computation.
To validate the effectiveness of the proposed feature graph learning, we employ it for 3D point cloud denoising with 3D coordinates and surface normals as features, leading to state-of-the-art performance.  


\end{document}